\documentclass[runningheads]{llncs}
\usepackage[dvipsnames]{xcolor}
\usepackage{graphicx}
\usepackage{afterpage}
\usepackage{romannum}
\usepackage{pgf-pie}
\usepackage{dirtytalk}
\usepackage{subcaption}
\usepackage{mdframed}
\usepackage{booktabs}
\usepackage[breaklinks,colorlinks]{hyperref}
\usepackage{multirow}
\usepackage{lscape}
\usepackage[first=0,last=9]{lcg}
\usepackage{color, colortbl}
\usepackage{notoccite}
\usepackage{cite}

\definecolor{Gray}{gray}{0.9}

\newcommand\crule[3][black]{\textcolor{#1}{\rule{#2}{#3}}}

\begin{document}

\title{BaDLAD: A Large Multi-Domain Bengali Document Layout Analysis Dataset}
%
\author{\small{
Md. Istiak Hossain Shihab$^{\star,1}$,
Md. Rakibul Hasan$^{\star,2}$, Mahfuzur Rahman Emon$^{\star,2}$, Syed Mobassir Hossen$^{1}$, Md. Nazmuddoha Ansary$^{1}$, Intesur Ahmed$^{1,4}$, Fazle Rabbi Rakib$^{1,2}$, Shahriar Elahi Dhruvo$^{1,2}$, Souhardya Saha Dip$^{1,2}$, Akib Hasan Pavel$^{1}$, Marsia Haque Meghla$^{1}$, Md. Rezwanul Haque$^{1}$, Sayma Sultana Chowdhury$^{2}$, Farig Sadeque$^{1,3}$,
Tahsin Reasat$^{1,4}$,
Ahmed Imtiaz Humayun$^{\dagger,1,5}$,
Asif Sushmit$^{\dagger,1,6}$}}
\authorrunning{M. Istiak et al.}

\pagestyle{empty}
%

\institute{
$^1$Bengali.AI, $^2$Shahjalal University of Science and Technology, $^3$BRAC University, $^4$Vanderbilt University, $^5$Rice University, $^6$RPI} 

\maketitle 
\begin{abstract}

While strides have been made in deep learning based Bengali Optical Character Recognition (OCR) in the past decade, absence of large Document Layout Analysis (DLA) datasets has hindered the application of OCR in document transcription, e.g., transcribing historical documents and newspapers. Moreover, rule-based DLA systems that are currently being employed in practice are not robust to domain variations and out-of-distribution layouts. To this end, we present the first multi-domain large \textbf{B}eng\textbf{a}li \textbf{D}ocument \textbf{L}ayout \textbf{A}nalysis \textbf{D}ataset: \textbf{BaDLAD}.
This dataset contains \textit{$33,695$ human annotated document samples from six domains} - i) books and magazines ii) public domain govt. documents iii) liberation war documents iv) new newspapers v) historical newspapers and vi) property deeds; with $710K$ polygon annotations for four unit types: text-box, paragraph, image, and table.
Through preliminary experiments benchmarking the performance of existing state-of-the-art deep learning architectures for English DLA, we demonstrate the efficacy of our dataset in training deep learning based Bengali document digitization models.  
\let\thefootnote\relax\footnotetext{Symbols $\star$ and $\dagger$ denote equal contribution.}
\let\thefootnote\relax\footnotetext{Project website: \url{https://bengaliai.github.io/badlad}}


\keywords{Handwritten Document Images \and Layout Analysis (Physical and Logical) \and Mobile/Camera-Based \and Other Domains \and Typeset Document Images}

\end{abstract}
\section{Introduction}
Understanding the layout of amorphous digital documents is a crucial step in parsing documents into organized machine-readable formats that are usable in real-world applications. Despite tremendous developments in machine learning (ML) methods and deep neural networks (DNNs) in recent decades, transcription of documents, e.g., historical books, remains a difficult challenge~\cite{10.1145/3534678.3539043}. Document layout analysis (DLA) is a preprocessing phase of a document transcription pipeline that detects and parses the structure of a document~\cite{binmakhashen2019document} by segmenting it into semantic units such as paragraphs, text-boxes, images and tables. Such segmented units are then transcribed via Optical Character Recognition (OCR) methods, for which robust algorithms have been proposed in literature~\cite{ocr,ocrsota}. The preprocessing step performed by DLA systems is often challenging due to different factors, e.g., free-writing style, deteriorating and faded text, ink spilling, and artistic lettering. Antiquated property documents, stained and torn papers and vague handwritten scripts make this task even more difficult \cite{binmakhashen2019document}. Robust DLA methods are therefore a major requirement for the digitization of handwritten records.

A DLA pipeline comprises a number of steps that may differ among approaches based on the layout of the specific document category and analysis goals~\cite{binmakhashen2019document}. Although rule-based algorithms and heuristic approaches were the standard for DLA in its earlier days~\cite{ahmad2016information}, recent decades have seen a major push towards solutions that use object detection models. Especially with the inception of DNNs, the accuracy and speed of such frameworks have greatly improved~\cite{girshick2015fast,ren2015faster,he2017mask,https://doi.org/10.48550/arxiv.2005.12872} paving the way for DNN based DLA methods~\cite{10.1145/3534678.3539043}. While datasets like \textit{DocBank}~\cite{li2020docbank} and \textit{PubLayNet}~\cite{zhong2019publaynet} are large enough to cater to the sample complexity of DNN based DLA frameworks, the datasets lack diversity in the orientation of annotations - which are mostly axes aligned. Moreover, such datasets contain data from a single domain, e.g., pdf articles from PubMed for \textit{PubLayNet}. Therefore, DNNs trained on such homogeneous sources, risk being vulnerable towards domain or distribution shifts~\cite{domainAdapt2019}.

In this paper, we present a dataset of documents collected from the wild, from multiple domains containing text with diverse layouts and orientations. Our dataset is the first large scale multi-domain document layout analysis dataset for Bengali. Our main contributions are as follows:
\begin{itemize}

    \item  We present a human-annotated dataset of 33,693 documents collected in the wild ~\say{\textbf{BaDLAD}}, for document layout analysis in Bengali. {BaDLAD is the largest organic dataset for Bengali DLA to the best of our knowledge.} Our dataset contains 710K polygon annotations for four unit/segment types: i) text-box, ii) paragraph, iii) images, and iv) table.
    
    \item BaDLAD comprises data collected from six different domains, i) books and magazines, ii) public domain govt. documents, iii) liberation war documents, iv) new newspapers, v) historical newspapers, and vi) property deeds. {To the best of our knowledge, BaDLAD is also the first multi-domain DLA dataset for Bengali.}
    
    \item We present preliminary results benchmarking the performance of popular DNN based DLA methods on BaDLAD. {We show that existing English DLA state-of-the-art models, fine-tuned on BaDLAD, exhibit improved performance on Bengali document layout analysis tasks in the multi-domain setting.}
    
\end{itemize}

Apart from this, we also present an additional \textit {4 million} un-annotated images including captured, scanned and printed documents that can be used for unsupervised DLA. The following sections are organized as follows. In Sec.~\ref{sec:related} we discuss related work on Document Layout Analysis that is present in literature. In Sec.~\ref{sec:challenges} we discuss the challenges present in Bengali DLA, also motivating the need for documents collected from the wild. In Sec.~\ref{sec:badlad} we present discussions on our collection protocols, annotation pipeline and statistics of our collected dataset. In Sec.~\ref{sec:bench} we present preliminary benchmarks on our dataset and following that in Sec.~\ref{sec:conclusion} we present conclusions and future directions. We make the codes for our benchmarking models and the corresponding data analysis publicly available under the CC BY-SA 4.0 license.


\section{Related Work}
\label{sec:related}
\textbf{Document Layout Analysis.} According to~\cite{zhong2019publaynet}, Zhong et al. generated and distributed the \textit{PubLayNet} dataset for document layout analysis, which includes automatically annotated data through matching with XML representations. Using an implementation of the Detectron algorithm, they trained an F-RCNN model and an M-RCNN model using PubLayNet. This dataset is claimed to be the largest one out there, containing 1 million pdf pictures of PMCOA (PubMed Centre Open Access) articles. PubLayNet data represent only scientific papers, which is topic-specialized and reduces layout diversification.

Li et al. presented a dataset \textit{DocBank} which contains 500K document-level images in English with fine-grained token-level annotations for structure analysis. They performed experiments on this dataset using four baseline models (BERT, RoBERT, LayoutLM, and Faster R-CNN) and claimed that the dataset can be utilized in any sequence labeling model~\cite{li2020docbank}. However, this dataset is based on automatically annotated English documents, which hurts its generalizability.

Pfitzmann et al. presented a manually annotated document layout dataset \textit{DocLayNet} in COCO format containing data from diverse sources~\cite{10.1145/3534678.3539043}. They presented benchmark accuracies for a collection of standard object detection models (MASK R-CNN, Faster R-CNN and YOLOv5) and analyzed models trained on PubLayNet, DocBank, and DocLayNet \cite{10.1145/3534678.3539043}.
Non-overlapping, vertically oriented, rectangular boxes were permitted during the annotation process.
According to them, human-annotated datasets provide more credible layout ground truth on a diverse range of publication and typesetting styles compared to DocBank and PubLayNet.
 Oliveira et al. \cite{oliveira2017fast} proposed a block based classification method to detect the layout of structured image documents rapidly and automatically through one dimensional CNN approach with a bi-dimensional CNN to compare performance and demonstrate their work. \\\\
\textbf{Bengali Document Layout Analysis.} Clausner et al.~\cite{8270161} presented several methods including four open-source SOTA systems for the evaluation of page analysis and identification algorithms for ancient manuscripts written in Bengali through their comparative assessment on this topic. This dataset is available in ICDAR challenges.
For SOTA methods, they used Tesseract 3.04 and 4.0 with internal binarization and long short-term memory units (LSTMs). 
Bangla OCR-I used Google’s Tesseract OCR engine for text classification and only works on printed scripts, whereas Bangla OCR-II's primary classification engine is a feature-based SVM and cannot handle intricate frames~\cite{8270161}.  
Some current datasets for Bengali document layout analysis are already being utilized in document processing tasks, although their size is rather limited~\cite{zhong2019publaynet}.
\section{Challenges of Bengali Document Layout Analysis}
\label{sec:challenges}

Bengali, one of the most widely spoken languages globally, is characterized by a large number of native speakers, estimated at almost 300 million, with 37 million international speakers. 
Despite its extensive usage, the field of Document Language Analysis (DLA) in Bengali remains in its nascent stages, with limited research and resources available on the subject. The synthetic data generation approach, commonly adopted by well-known datasets such as PubLayNet and DocBank, does not apply to Bengali given the majority of the publicly accessible Bengali documents are either scanned images or captured photographs of the original document and thus cannot be annotated using automatic algorithms. Moreover, such datasets
are comprised of synthetic, born-digital documents and are carefully curated, resulting in annotations with exclusively horizontal and vertical boundaries. In contrast, our dataset incorporates irregularly-shaped polygon annotations and preserves their original boundaries. It is our belief that this approach will enhance the precision of layout detection and related challenges, such as optical character recognition and form detection.

The Bengali script, being a non-Latin-based script, possesses another challenge for DLA tasks. Bengali has an intricate writing system encompassing inflections, multiple script forms, and character composites. This is because individual characters can exhibit different forms based on their position within a word or the preceding and succeeding letters. Furthermore, certain characters in Bengali may be represented through a combination of multiple characters \cite{grapheme2021}, which presents a challenge for models to identify them accurately. These complexities can result in inaccuracies in layout analysis, as the models may not be capable of discerning between the text elements and the interconnections among them.

The historical nature of printed Bengali documents, dating back to the early 1800s, coupled with the prevalence of typographical variations and the printing styles of ancient literature present significant difficulties for the document layout analysis (DLA) task in this language.
Additionally, the complexity of the layout, frequently unintelligible handwriting, deteriorating paper quality, and non-standard formatting of modern Bengali legal documents further exacerbate the state of this area of research. 
Given the recent advances in massively data-driven deep learning techniques, development of a machine-trainable, hand-annotated dataset with sufficient diversity to address these challenges should be a priority-- which is precisely what is proposed in the present paper.

\section{\textit{B}eng\textit{a}li \textit{D}ocument \textit{L}ayout \textit{A}nalysis \textit{D}ataset: \textit{BaDLAD}}
\label{3}
\label{sec:badlad}

BaDLAD comprises of data collected from six different domains. The dataset contains annotations for four semantic unit types via polygon annotations. In this section we first provide descriptions for the selected data domains and justifications for the semantic unit types. Following that we discuss our annotation pipeline and statistics of the collected data.

\subsection{Semantic Units for Layout Segmentation}
\label{subsec:units}

We started by scraping $\sim20,000$ Bengali PDF files from publicly available online repositories for books. To explore the layout diversity of these books
, we trained a self-supervised SwAV~\cite{caron2020unsupervised} model which generates prototypes that can be considered as the cluster centers of the model's embedding space. Upon inspecting the cluster centers, and manually inspecting a number of representatives from each cluster, we noticed four major semantic categories in which the layout can be partitioned:

\begin{itemize}

    \item[\textbullet] \textbf{Text-box} : A small isolated collection of letters, numbers, word or group of words, e.g., page number, book name, chapter name, headline/ title, or incomplete non-contiguous sentences.
    
    \item[\textbullet] \textbf{Paragraph} : A collection of text that is made up of one or more sentences and deals with a single topic or idea and is separated from other paragraphs by a line break or indentation. A single word can be considered as a paragraph when it is in context and makes a meaningful point or statement on its own, e.g., in a dialogue.
    
    \item[\textbullet] \textbf{Image}: Representation of any visual object that is not only comprised of text, e.g., logo, pictures, graphical handwritten signatures.
    
    \item[\textbullet] \textbf{Table} : Structured set of data made up of rows and columns, which may or may not have table headers or borders.
    
\end{itemize}

We did not find a significant number of list elements in the clusters. Hence we did not include the list category as a semantic unit in our dataset. In our dataset, we have annotated lists as a collection of text-boxes or paragraphs, depending on which of the aforementioned definitions the list elements are closest to. 

\afterpage{\clearpage}

\begin{figure}[h]
    \centering
    \begin{subfigure}{0.30\textwidth}
        \includegraphics[width=\linewidth]{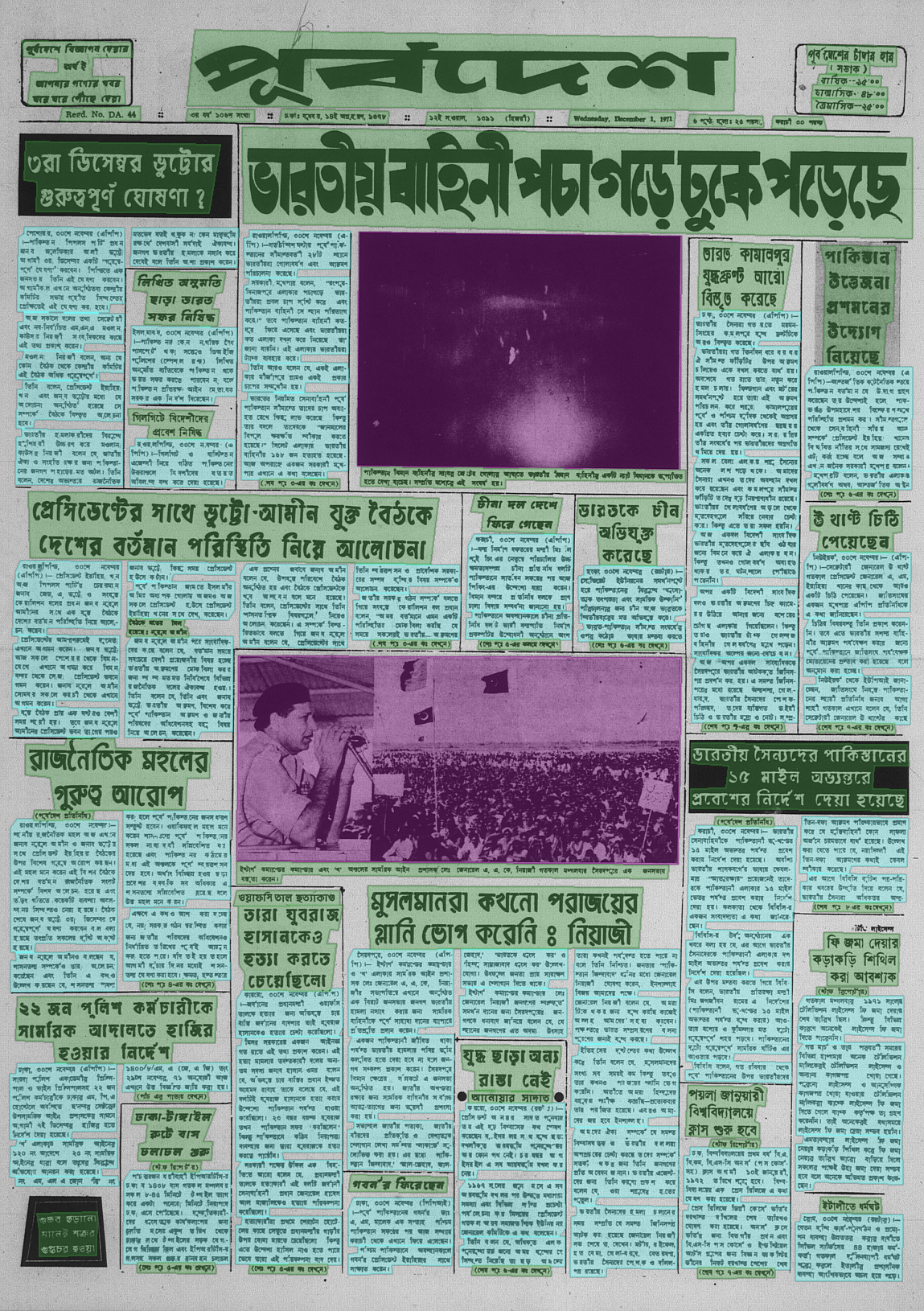}
        \caption{Newspaper (Hist.)}
        \label{fig:old_newspaper}
    \end{subfigure}
    \begin{subfigure}{0.30\textwidth}
        \includegraphics[width=\linewidth]{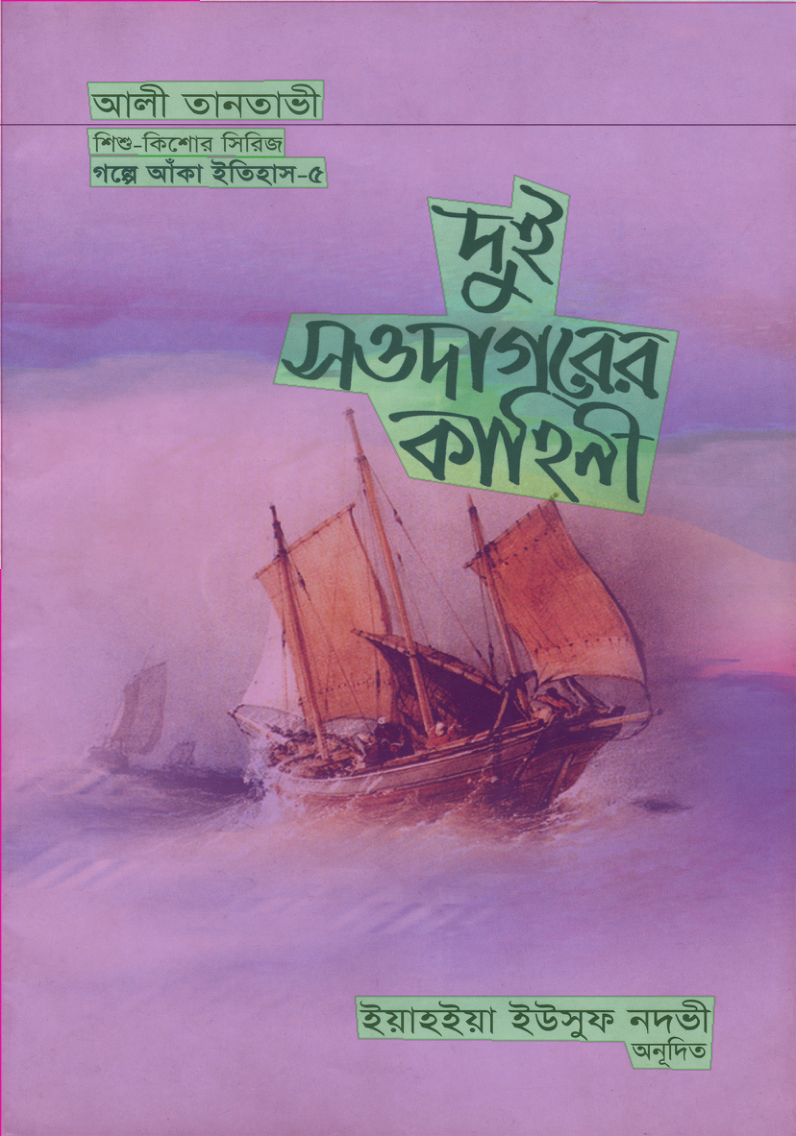}
        \caption{Book Cover}
        \label{fig:book_cover}
    \end{subfigure}
    \begin{subfigure}{0.33\textwidth}
        \includegraphics[width=\linewidth]{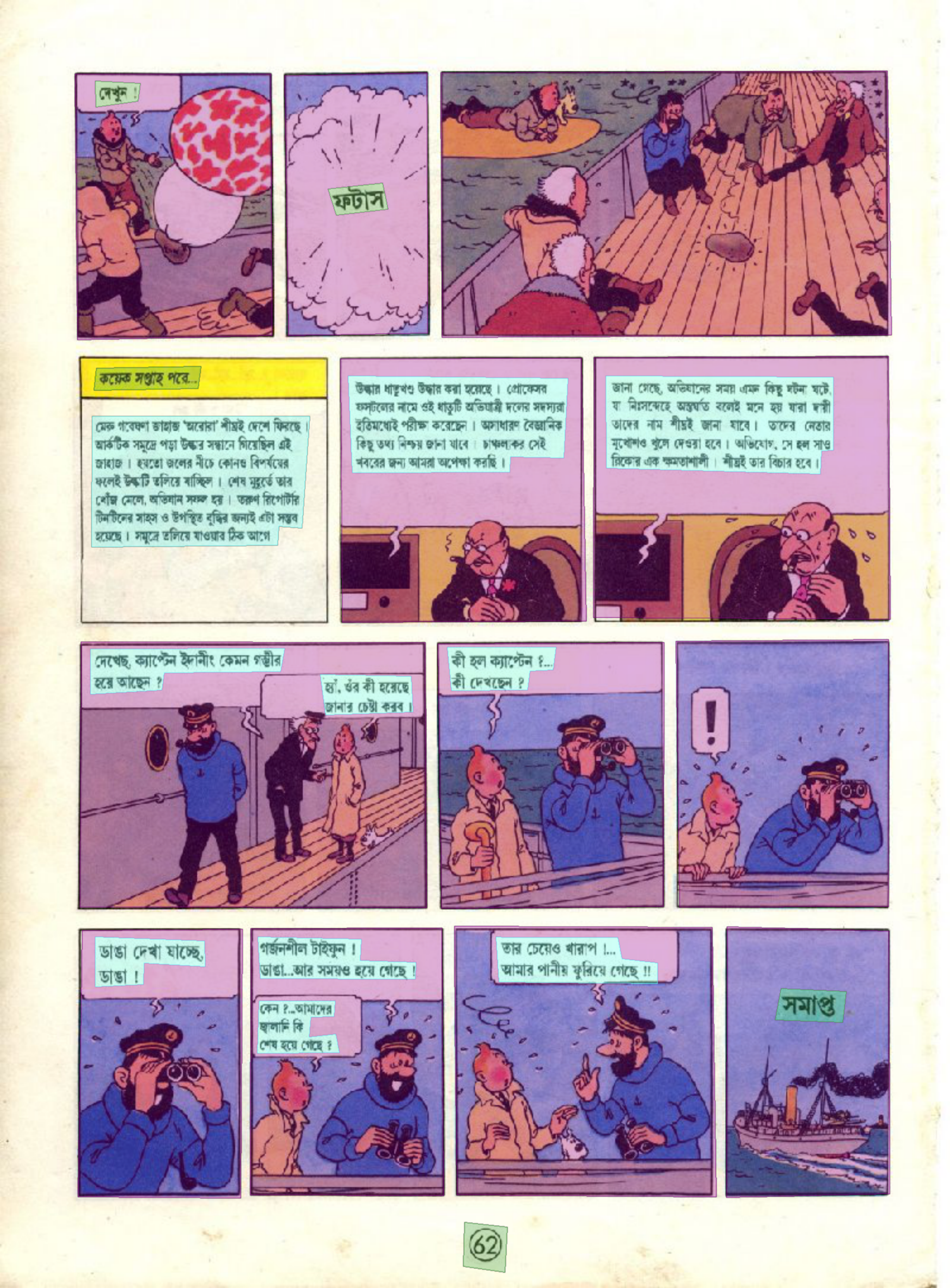}
        \caption{Comic}
        \label{fig:comic}
    \end{subfigure}

    \begin{subfigure}{0.30\textwidth}
        \includegraphics[width=\linewidth]{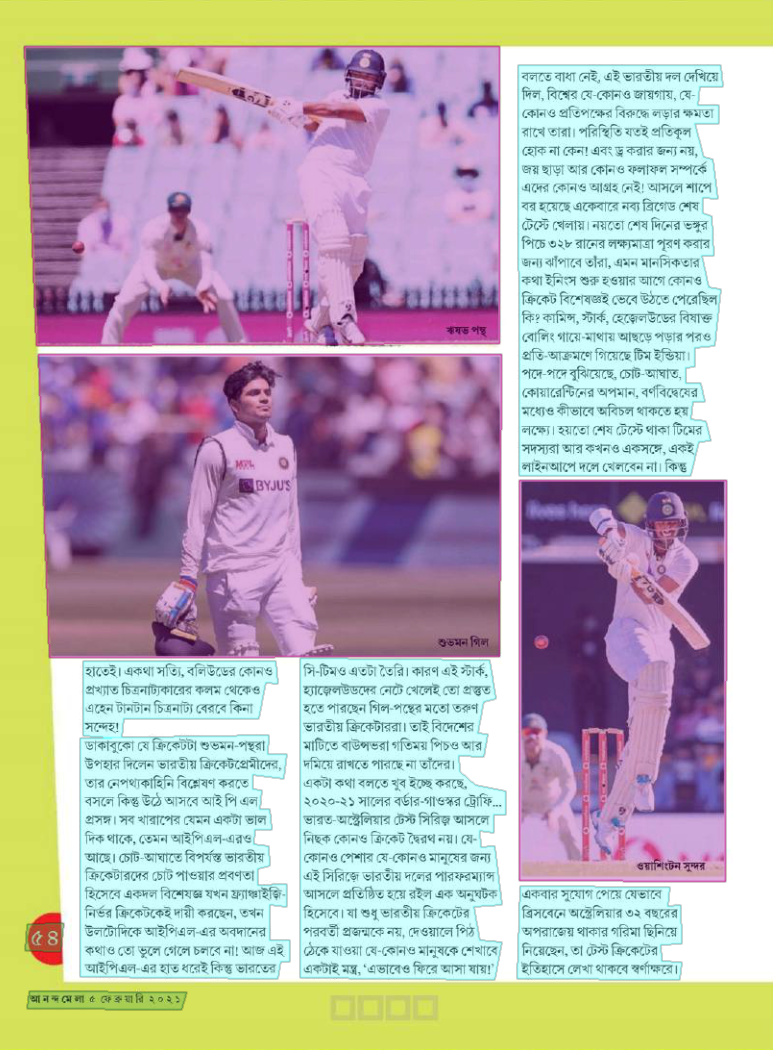}
        \caption{Magazine}
        \label{fig:magazines}
    \end{subfigure}
    \begin{subfigure}{0.33\textwidth}
        \includegraphics[width=\linewidth]{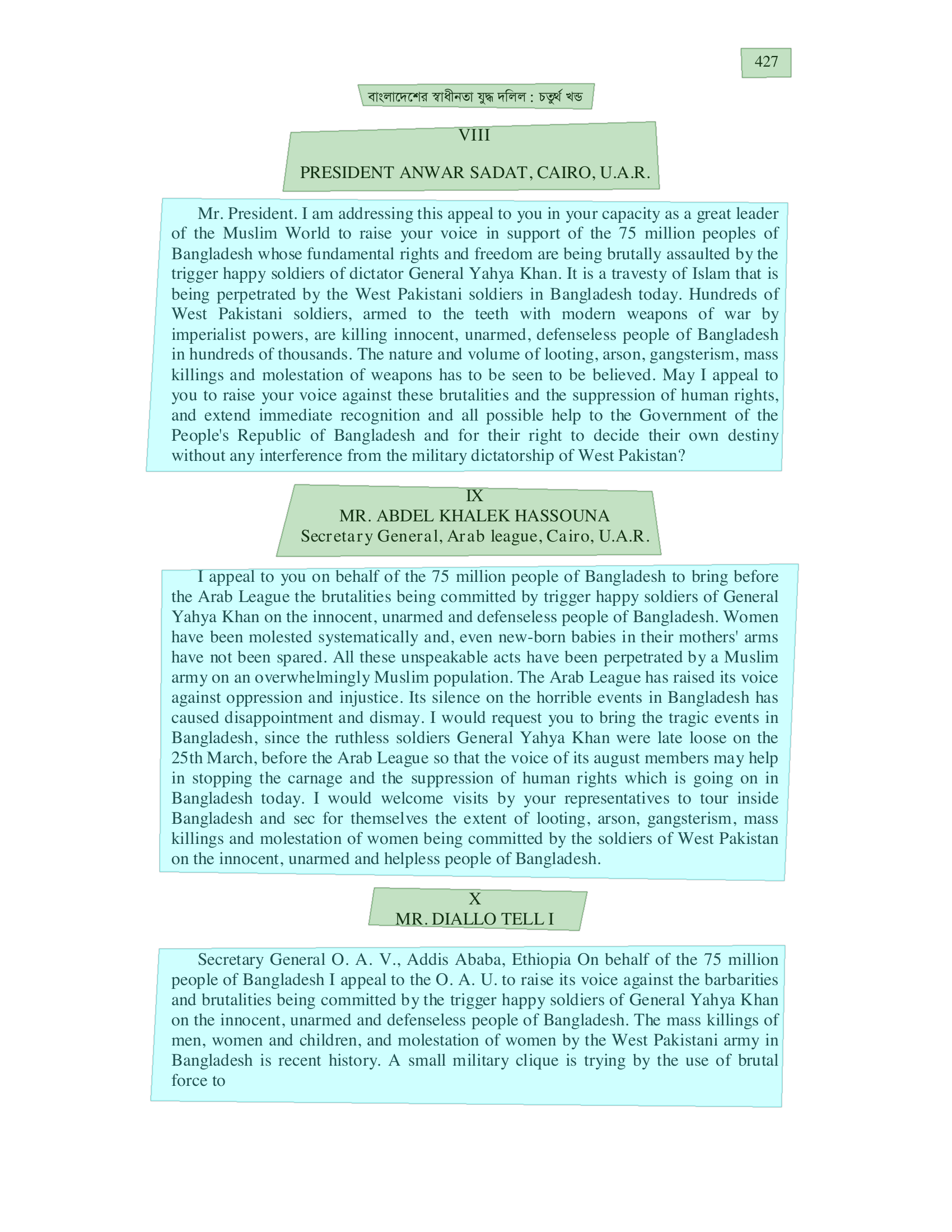}
        \caption{Liberation War Doc}
        \label{fig:lib_doc}
    \end{subfigure}
    \begin{subfigure}{0.32\textwidth}
        \includegraphics[width=\linewidth]{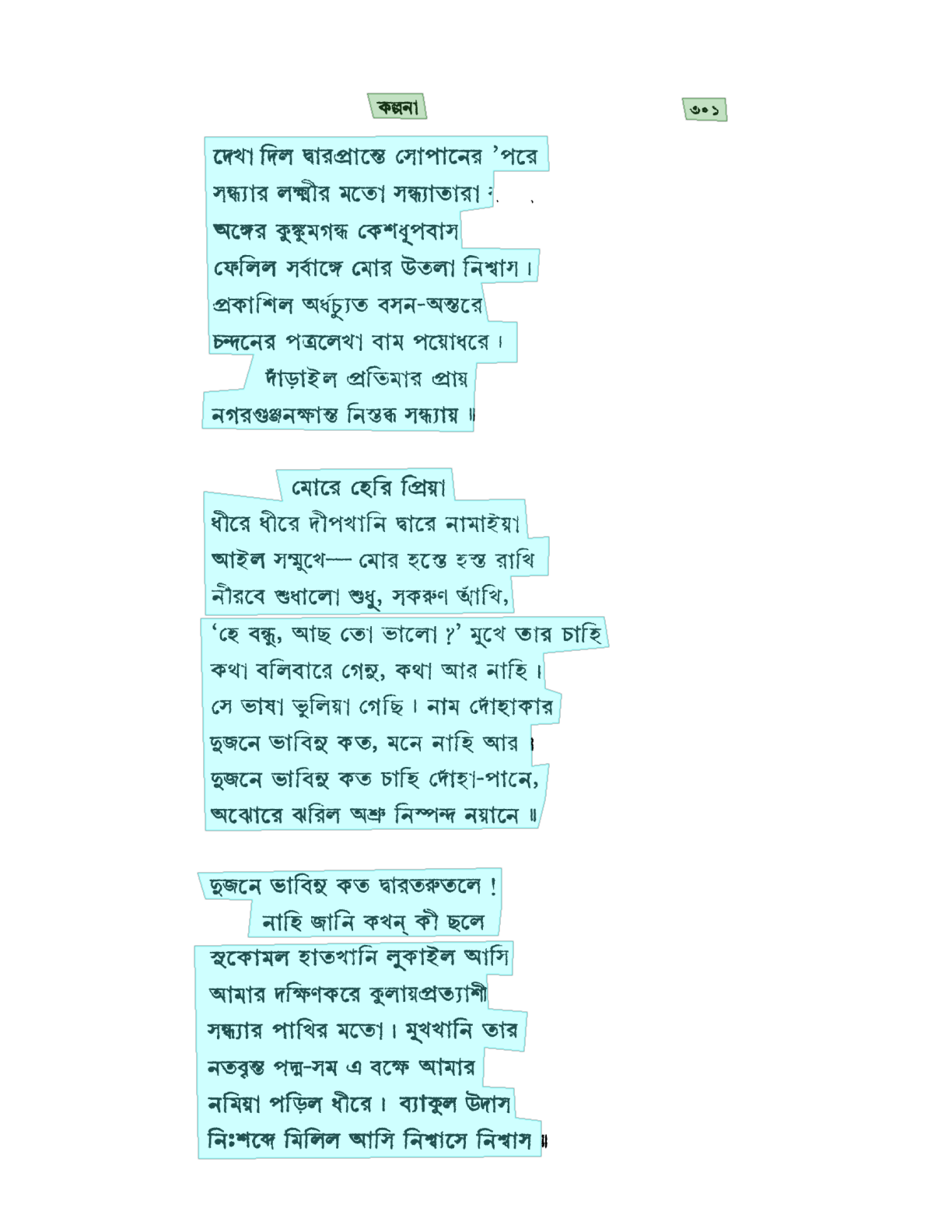}
        \caption{Poem}
        \label{fig:poems}
    \end{subfigure}
    
    \begin{subfigure}{0.31\textwidth}
        \includegraphics[width=\linewidth]{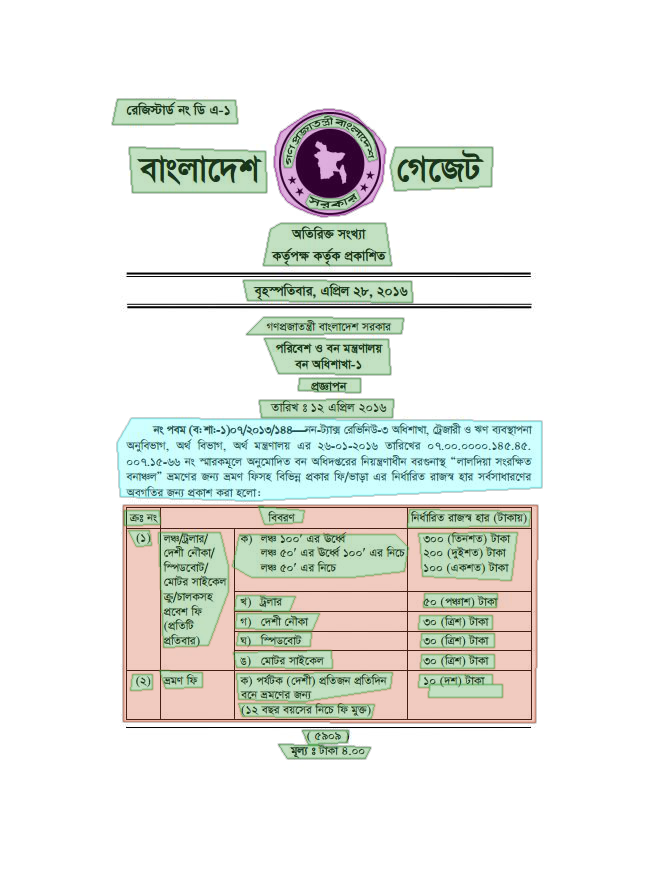}
        \caption{Govt Doc}
        \label{fig:gov_doc}
    \end{subfigure}
    \begin{subfigure}{0.31\textwidth}
        \includegraphics[width=\linewidth]{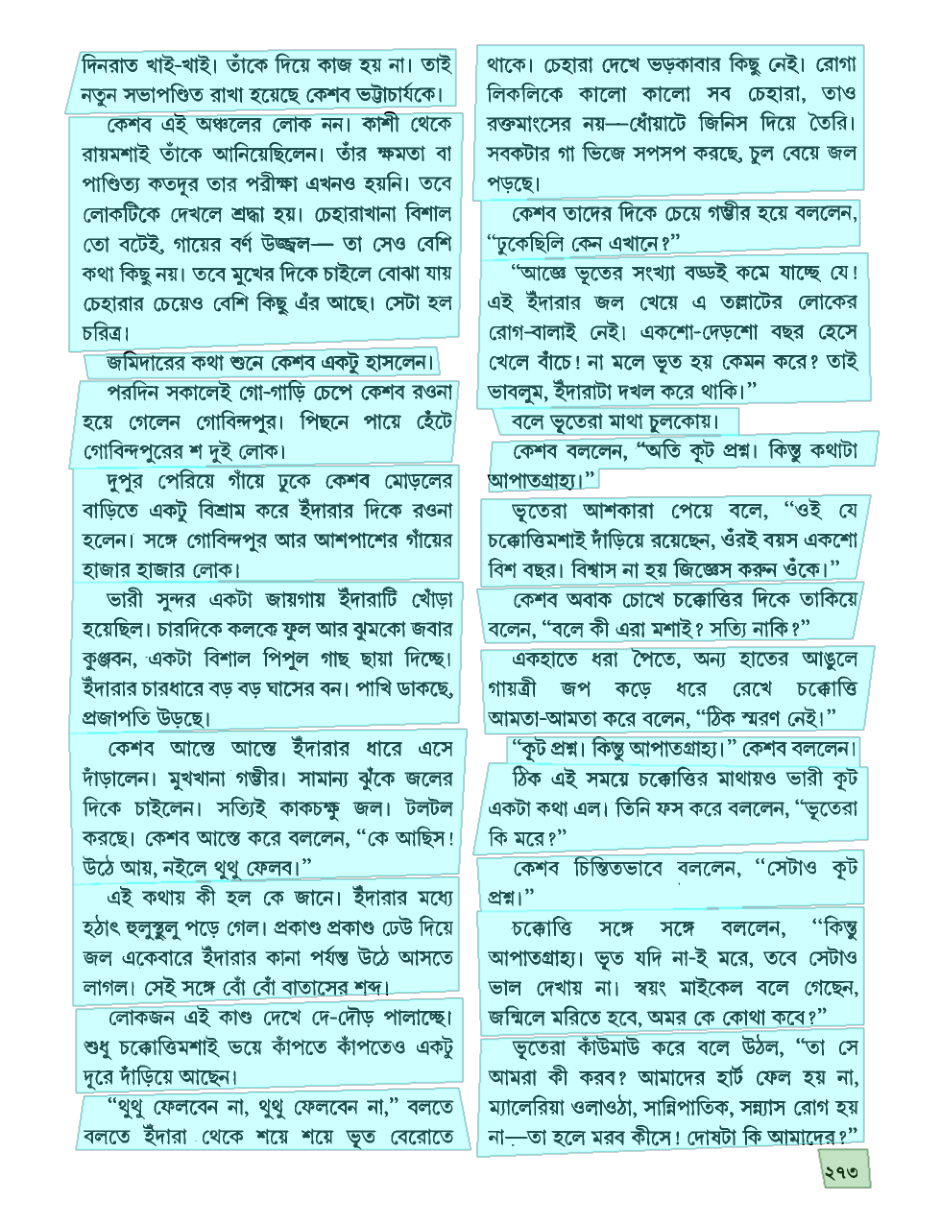}
        \caption{1-page-2-Column}
        \label{fig:sin_page_dub}
    \end{subfigure}
    \begin{subfigure}{0.33\textwidth}
        \includegraphics[width=\linewidth]{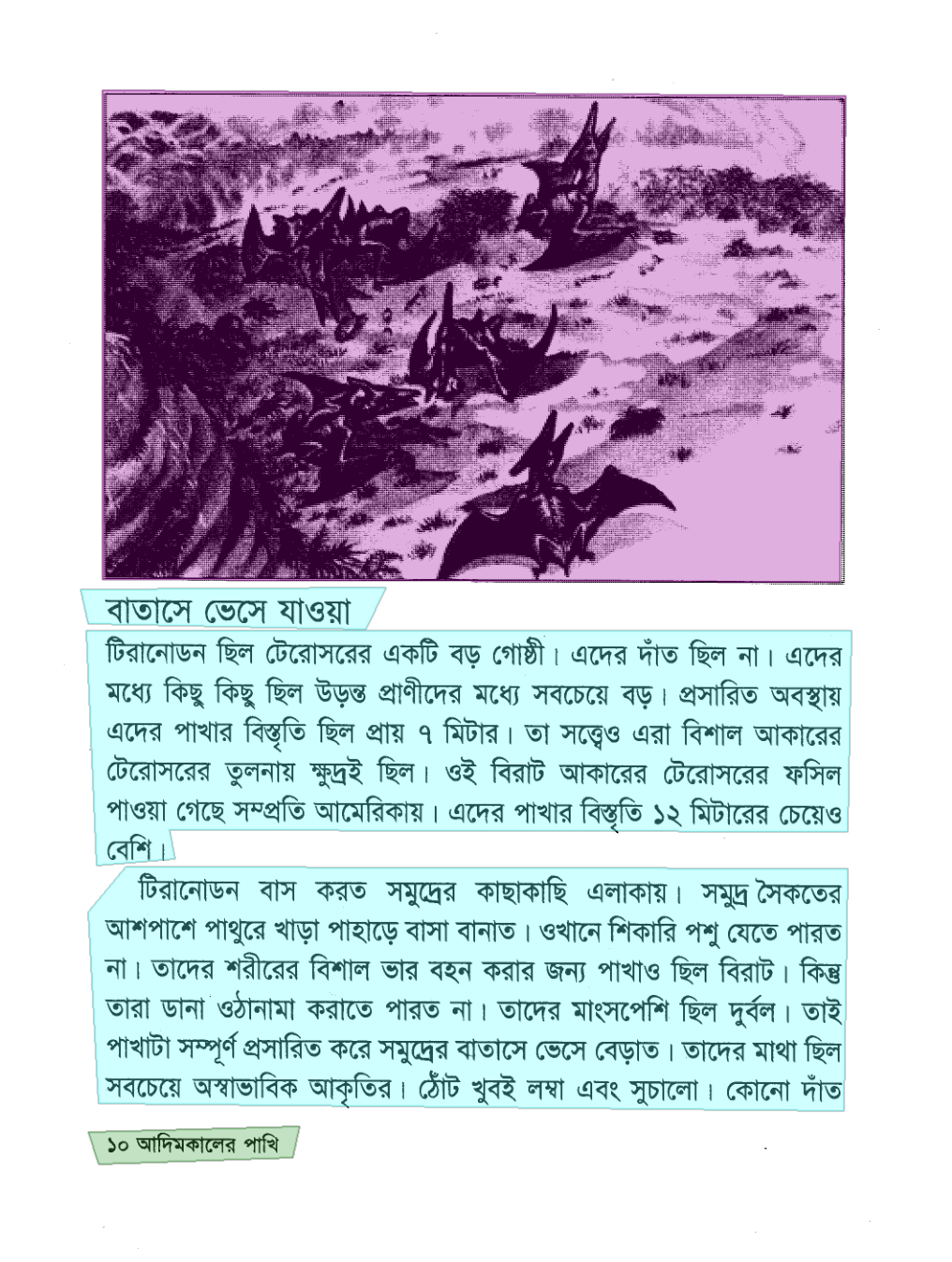}
        \caption{1-page-1-Column}
        \label{fig:sin_page_sin}
    \end{subfigure}
    
    \begin{subfigure}{0.32\textwidth}
        \includegraphics[width=\linewidth]{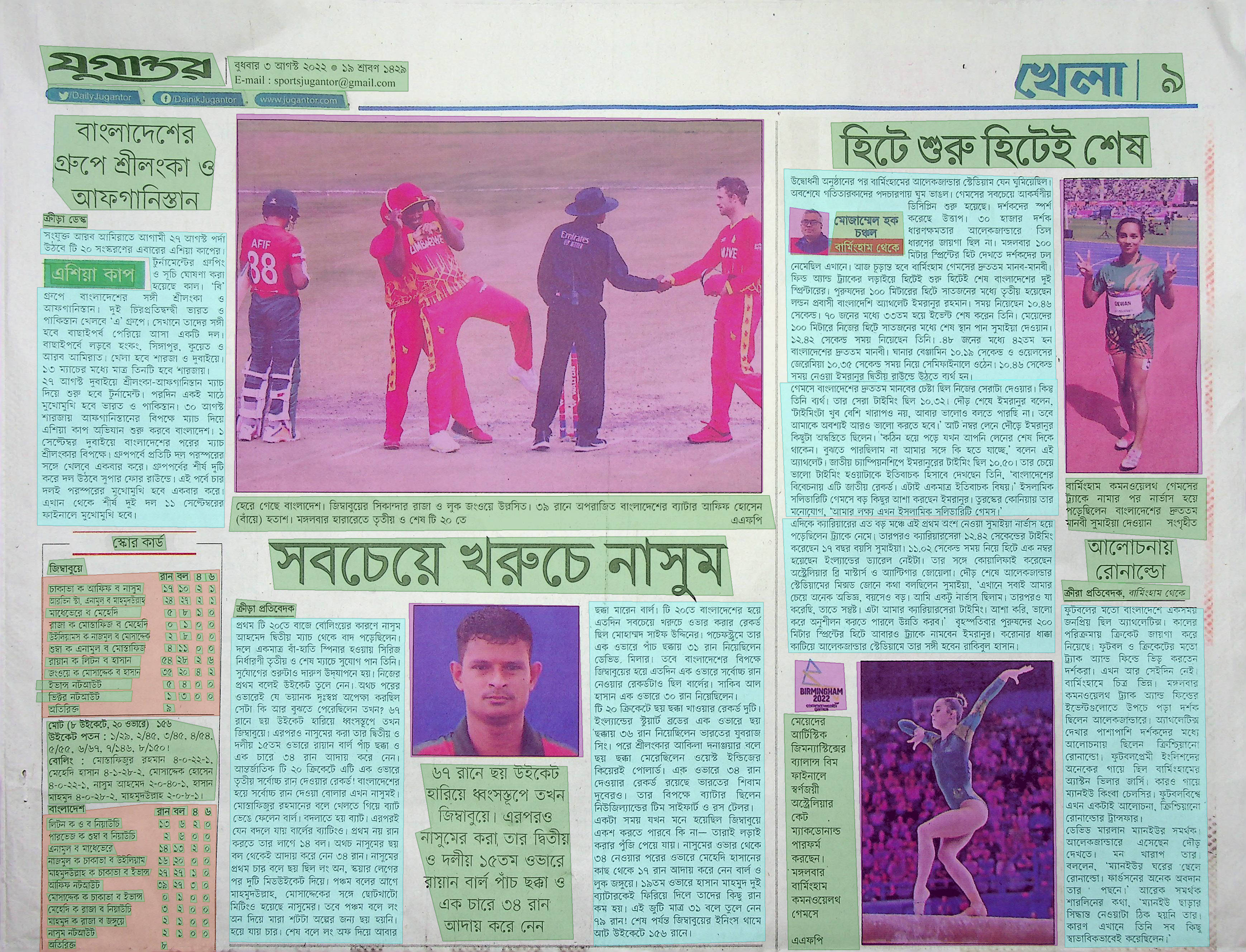}
        \caption{Newspaper(New)}
        \label{fig:new_newspaper}
    \end{subfigure}
    \begin{subfigure}{0.32\textwidth}
        \includegraphics[width=\linewidth]{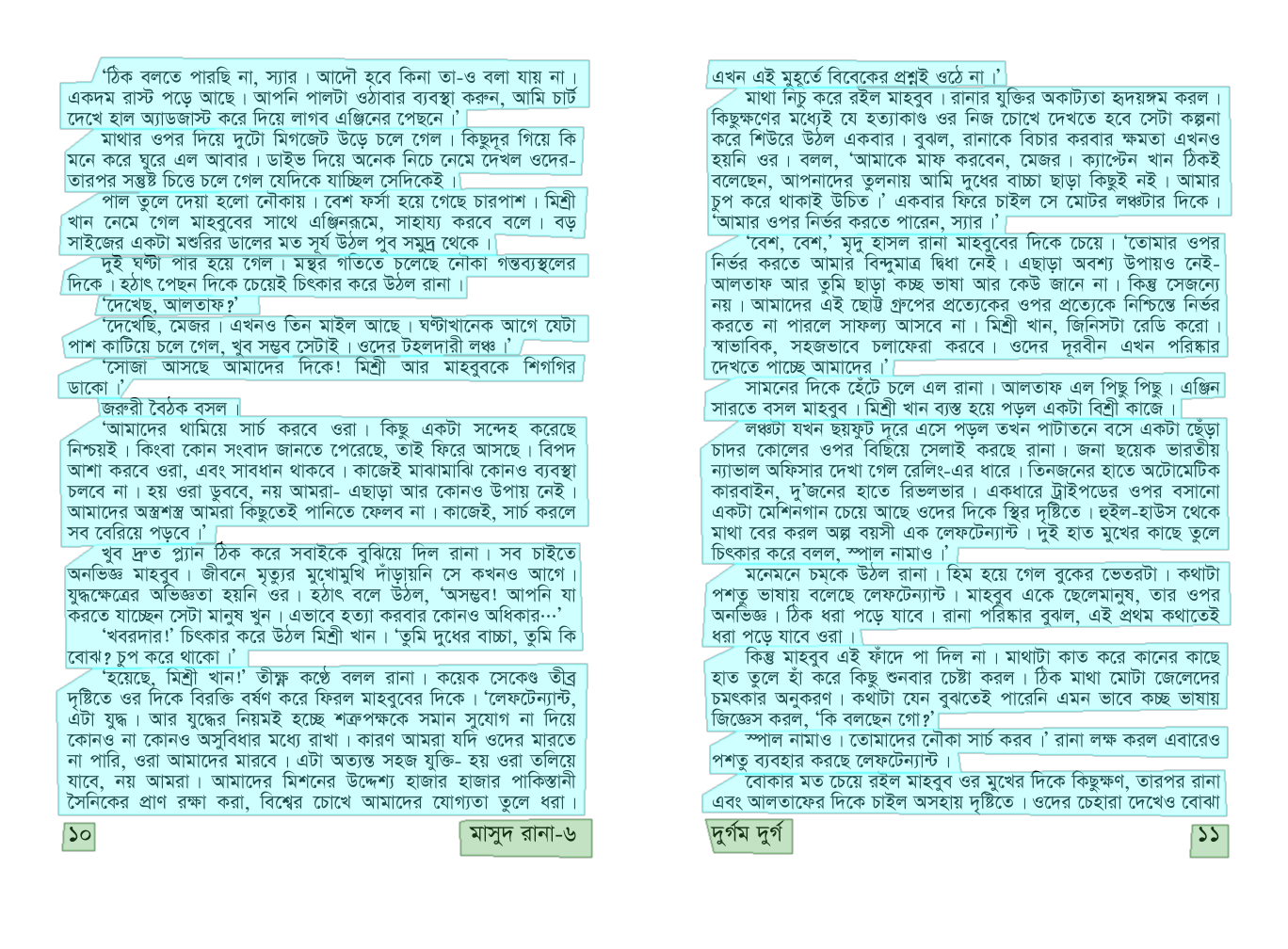}
        \caption{2-page-1-Scan}
        \label{fig:dub_page}
    \end{subfigure}
    \begin{subfigure}{0.32\textwidth}
        \includegraphics[width=\linewidth]{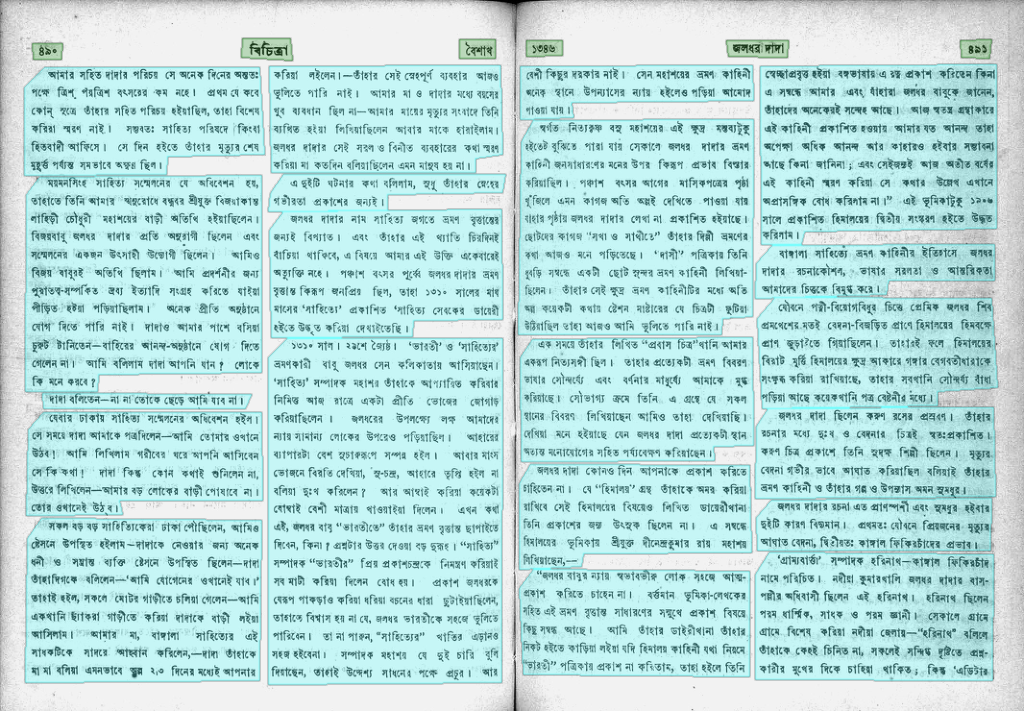}
        \caption{2-page-2-column-1-scan}
        \label{fig:dub_page_dub}
    \end{subfigure}
\caption{Different layout categories present in the BaDLAD dataset. Annotations are color coded as: ~\crule[YellowGreen]{8pt}{8pt} Text-box,~\crule[Aquamarine]{8pt}{8pt} Paragraph,~\crule[Thistle]{8pt}{8pt} Image,~\crule[Melon]{8pt}{8pt} Table. We do not present examples from the \textit{Property Deeds} domain to ensure confidentiality.}
\label{fig:exampleSamples}
\end{figure}

\subsection{Domain Categories and Sources}
\label{section:distribution}

To make the dataset diverse and complex, we collected documents from a wide range of domains, e.g., Novels, Magazines, Poems, Newspapers, Government Documents, Property Deeds, Liberation War Documents, which we have binned into the following categories based on sources. We have also presented representative samples in Fig.~\ref{fig:exampleSamples}.\\

\noindent
\textbf{Magazine and Books.} This domain comprises of samples from $\sim20,000$ Bengali PDFs scraped from publicly available online repositories, as mentioned in Sec.~\ref{subsec:units}.  The collection comprised of books, magazines (fig.~\ref{fig:magazines}), poems (fig.~\ref{fig:poems}) and comics (fig.\ref{fig:comic}) with a very diverse set of layouts. We also take into account the book covers while sampling from the collected PDFs. All of the PDFs are scanned or photo captured versions of the original document, without any digital transcription. Literary works comprising mostly of text, e.g., novels, contain three major layout types - single page single column (fig. ~\ref{fig:sin_page_sin}), single page double column (fig.~\ref{fig:sin_page_dub}) and double page single scan (fig.~\ref{fig:dub_page} and ~\ref{fig:dub_page_dub}).

\noindent
\textbf{Historical Newspapers.} Historical Newspapers that have been published before December 1971 that were manually scanned. The typesetting of such newspapers are significantly different from new newspapers, e.g., in terms of font size, font style, glyphs of consonant conjuncts (fig.~\ref{fig:old_newspaper}).

\noindent
\textbf{New Newspapers.} Recently published newspapers manually captured by scanners and cameras (fig.~\ref{fig:new_newspaper}).

\noindent
\textbf{Liberation War Documents.} Taken from a $15$-part collection of liberation war documents, manually scanned (fig.~\ref{fig:lib_doc}).

\noindent
\textbf{Government Documents.} We have collected publicly available government documents by scraping from online repositories and by manually collecting and scanning. These documents comprise of both handwritten and printed characters along with logos, seals, tables, headers and graphical elements (fig.~\ref{fig:gov_doc}).

\noindent
\textbf{Property Deeds.} Confidential documents collected with consent via social media crowd-sourcing campaigns. We have anonymized the documents by removing sensitive and identifiable information and include them only in the hidden test dataset. These documents generally contain a lot of handwritten notes, signatures and free-form text, posing a challenging DLA task.

\subsection{Annotation and Validation}
\label{annotation}
\begin{figure}[t]
    \centering
    \includegraphics[width=\textwidth]{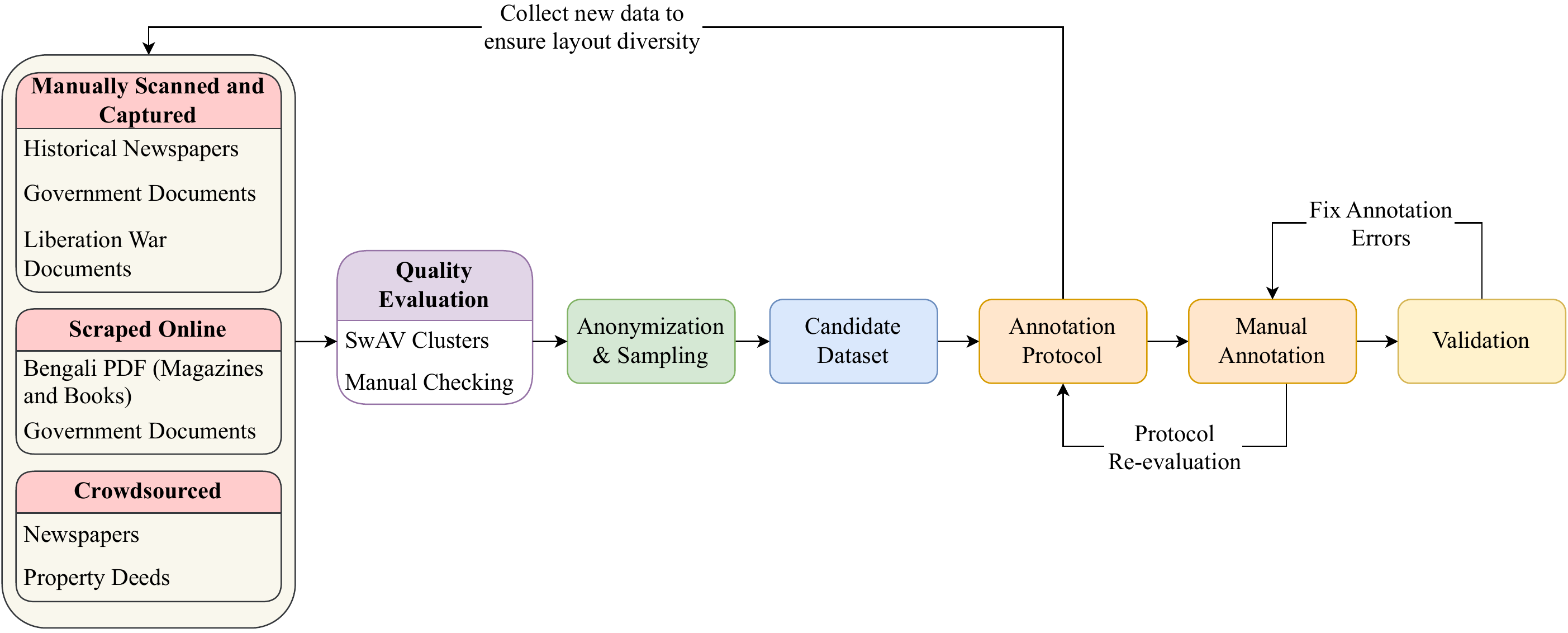}
    \caption{Data collection and annotation pipeline for the BaDLAD dataset. Candidate samples from the un-annotated dataset are collected and curated dynamically with annotation and validation tasks to ensure layout diversity and quality of images.}
    \label{fig:annotation_pipeline}
\end{figure}

To ensure diversity of samples, we chose 2 pages randomly from each scanned document, since pages from the same document have higher probability of being similar. A team of 13 annotators were trained to annotate document layout on the \say{Labelbox} platform. 
Polygon labeling was used because of the complex orientation of texts and images in our dataset (as can also be seen in Fig.~\ref{fig:exampleSamples}). 
Each annotator was tasked to segment all the semantic units in a given sample. We also kept track of the time required to annotate each sample as metadata, which can be considered as a segmentation hardness measure for each sample. 
The annotators annotated 33,693 samples in total over a course of four months. During annotation, three curators were assigned the task of annotation verification and curation. Any document with wrong annotations were resent to the original annotator for correction. The annotation guidelines were also dynamically updated during this process. In Fig.~\ref{fig:annotation_pipeline} we provide a brief overview of our data collection, annotation and validation process. 


\noindent
\textbf{Brief annotation guideline.} In order to obtain more effective data, we developed objective guidelines for the annotators which applied in a domain-unit specific manner. All plots or graphs were considered as images. For samples with double page scans, if any portion of the content of one page went to another, then the divided portions were annotated separately. In the case of poetry, if there were extra white spaces between lines then the lines were considered separate paragraphs. Bullets or numbered lists were separately annotated as text-boxes for one line sentences and paragraphs in the case of multi-line sentences. Hand-written texts were considered as texts except for signatures, signatures were marked as images. If there were any extra notes (e.g. URLs, post-scripts) along with a paragraph, then the extra portion was annotated as a text-box.
Vertical lines, advertisements/links were marked as text-boxes. Tables were annotated as tables, but the contents were marked according to the definitions as images, text-boxes or paragraphs. If for a sample, the contents from the other side of a scanned page were visible due to transparency, the text from the opposite were ignored and only the main text from the correct side was annotated.

\subsection{BaDLAD Statistics}
\begin{figure}[!ht]
    \centering
    \includegraphics[width=\textwidth]{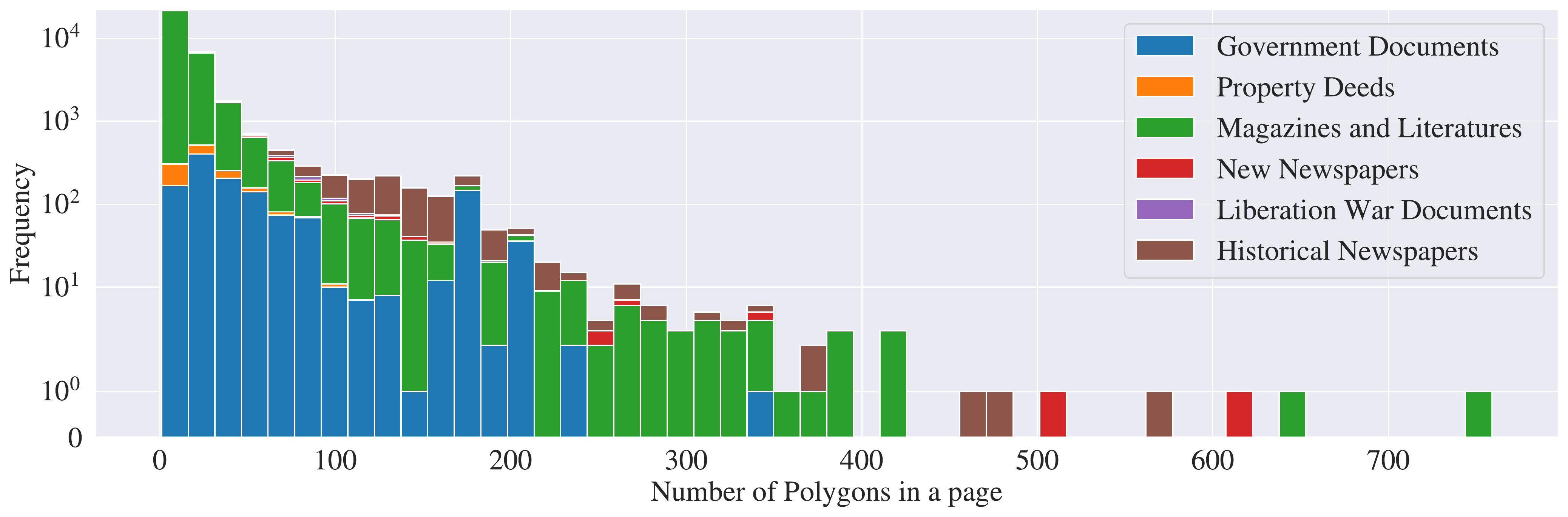}
    \caption{Histogram of Polygons per page stacked and colored by the domain presented in logarithmic scale. Samples from the \textit{Government Documents} domain contain a lower number of polygons in every page. Both \textit{Historical Newspapers} and \textit{New Newspapers} contains a higher number of polygons per page, which correlates with their higher avg. annotation time requirement according to Table.~\ref{tab:domainStat}. Samples from the \textit{Magazines and Books} domain contain a large diversity in number of polygons per page.}
    \label{fig:whole_ptype}
\end{figure}

After annotation and curation, the BaDLAD dataset comprises a total of 33,693 samples; of which $30054$ samples are from \textit{Magazines and Books}, $1285$ samples from \textit{Govt. Documents}, $1004$ samples from \textit{Liberation War Documents}, $861$ samples from \textit{Historical Newspapers}, $328$ samples from \textit{Property Deeds} and $161$ samples from \textit{New Newspapers}. While \textit{Magazines and Books} is the most prevalent domain, as discussed in Sec.\ref{section:distribution} and presented in Fig.~\ref{fig:exampleSamples}, the domain contains a large diversity of layouts from multiple sources. In Table.~\ref{tab:domainStat} and Table.~\ref{tab:domainStatTest}, we present the domain-wise number of annotations for every unit type, along with the time elapsed for annotation. We present it for the train and test splits separately as specified in Sec.~\ref{subsec:split}. If we consider the average time required to annotate a sample as a hardness measure for each sample, we can see that samples from the \textit{Historical Newspapers} and \textit{New Newspapers} domains are the most challenging. On the other hand, samples from the \textit{Liberation War Documents} domain is considerably easier, which can be attributed to the relatively larger volume of paragraph annotations. This distinction is also present in the number of polygon per page histogram, presented in Fig.~\ref{fig:whole_ptype}.

\begin{table}[htbp]
\centering
\resizebox{\textwidth}{!}{%
\begin{tabular}{@{}cccccccc@{}}
\toprule
\textbf{Domain} & \textbf{Samples} & \textbf{Text-box} & \textbf{Paragraph} & \textbf{Image} & \textbf{Table} & \textbf{\begin{tabular}[c]{@{}c@{}}Total Annotation\\ Time (Hours)\end{tabular}} & \textbf{\begin{tabular}[c]{@{}c@{}}Avg Annotation \\ Time (Minutes)\end{tabular}} \\ \midrule
Historical Newspapers & 516 & 25452 & 38990 & 1252 & 67 & 211.52 & 24.60 \\
New Newspapers & 96 & 3978 & 2507 & 494 & 36 & 27.35 & 17.10\\
Govt Documents & 771 & 44017 & 2260 & 762 & 514 & 78.44 & 6.10 \\
Magazines and Books & 18380 & 123099 & 162570 & 7734 & 594 & 948.58 & 3.10 \\
Property Deeds & 0 & 0 & 0 & 0 & 0 & 0.0 & 0.0\\
\begin{tabular}[c]{@{}c@{}}Lib War Docs\end{tabular} & 602 & 7330 & 3262 & 55 & 142 & 22.43 & 2.24 \\
\textbf{Total} & 20365 & 203876 & 209589 & 10297 & 1353 & 1288.33 & 3.80 \\ \bottomrule
\end{tabular}%
}
\caption{Domain-wise annotation statistics for BaDLAD (Train)}
\label{tab:domainStat}
\end{table}

\begin{table}[htbp]
\centering
\resizebox{\textwidth}{!}{%
\begin{tabular}{@{}cccccccc@{}}
\toprule
\textbf{Domain} & \textbf{Samples} & \textbf{Text-box} & \textbf{Paragraph} & \textbf{Image} & \textbf{Table} & \textbf{\begin{tabular}[c]{@{}c@{}}Total Annotation\\ Time (Hours)\end{tabular}} & \textbf{\begin{tabular}[c]{@{}c@{}}Avg Annotation \\ Time (Minutes)\end{tabular}} \\ \midrule
Historical Newspapers & 345 & 17611 & 26571 & 838 & 54 & 146.85 & 25.54 \\
New Newspapers & 65 & 3542 & 1902 & 237 & 24 & 22.57 & 20.83\\
Govt Documents & 514 & 27903 & 1497 & 482 & 301 & 52.63 & 6.14 \\
Magazines and Books & 11674 & 80581 & 103390 & 4949 & 376 &625.02 & 3.21 \\
Property Deeds & 328 & 6012 & 599 & 930 & 117 & 17.03 & 3.11 \\
\begin{tabular}[c]{@{}c@{}}Lib War Docs\end{tabular} & 402 & 4733 & 2370 & 16 & 104 & 16.06 & 2.40\\
\textbf{Total} & 13328 & 140382 & 136329 & 7452 & 976 & 880.17 & 3.96 \\ \bottomrule
\end{tabular}%
}
\caption{Domain-wise annotation statistics for BaDLAD (Test)}
\label{tab:domainStatTest}
\end{table}

\begin{figure}[htbp]
     \centering
     \begin{subfigure}[b]{.9\textwidth}
         \centering
         \includegraphics[width=\textwidth]{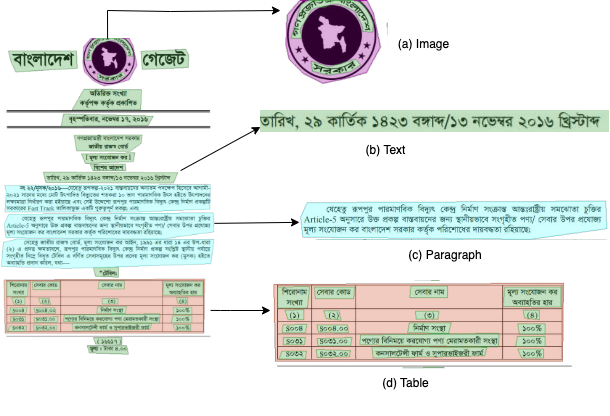}
         \caption{Semantic units for annotation}
         \label{categories}
     \end{subfigure}
          \begin{subfigure}[b]{.9\textwidth}
         \centering
         \includegraphics[width=\textwidth]{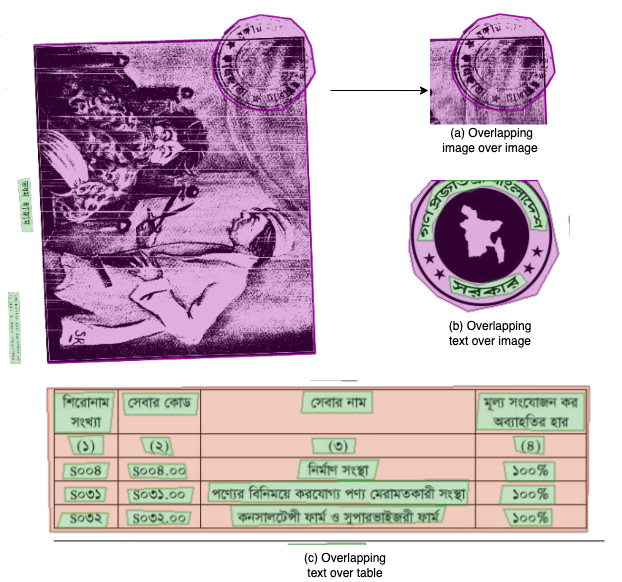}
         \caption{Overlapping annotations}
         \label{fig:Overlapping}
     \end{subfigure}
     \caption{Annotated samples from the BaDLAD dataset with semantic units (i). Annotation overlaps between different semantic units (ii).}
    \label{fig:Annotation_labels}
\end{figure}

\begin{figure}
    \includegraphics[width=\textwidth]{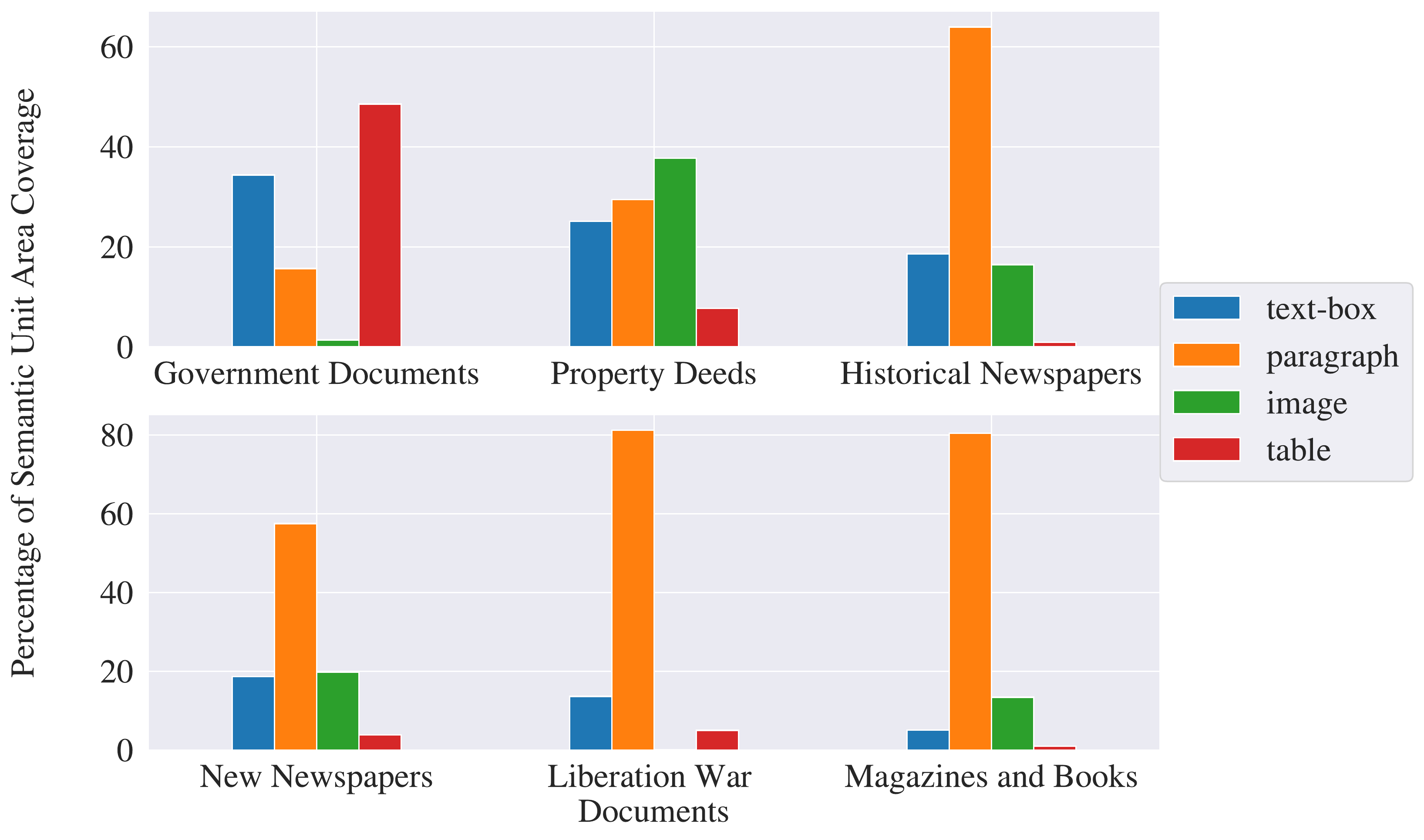}
    \caption{Area covered by polygons for every unit type, normalized by the total area of samples per domain. Except for \textit{Govt. Documents}, all the domains have a larger area covered by paragraphs. For the \textit{Govt. Documents} domain, even though the number of paragraphs is higher than the number of tables, the area covered by tables is significantly higher than that of paragraphs.}
    \label{fig:distribution_document_types}
\end{figure}

In Fig.~\ref{fig:distribution_document_types} we present the area covered by each unit type as a percentage of the total area of samples per domain, for the whole dataset. Here, area is calculated in pixel units. We can see that for Liberation War Documents, Magazines and Books and Newspapers, a large fraction of the images are covered in paragraphs. On the other hand, for Govt. Documents, a larger fraction of area is covered by tables. For all of these graphs, the percentage sum might be larger than $1$ since there can be significant overlaps in the area covered by two different annotations. For example, in Fig.~\ref{fig:Overlapping} we present examples of different overlaps which are frequently present in the dataset.
\renewcommand{\thesubfigure}{\roman{subfigure}}

\begin{figure}
    \centering
    \begin{subfigure}[b]{0.24\textwidth}
         \centering
         \includegraphics[width=\textwidth]{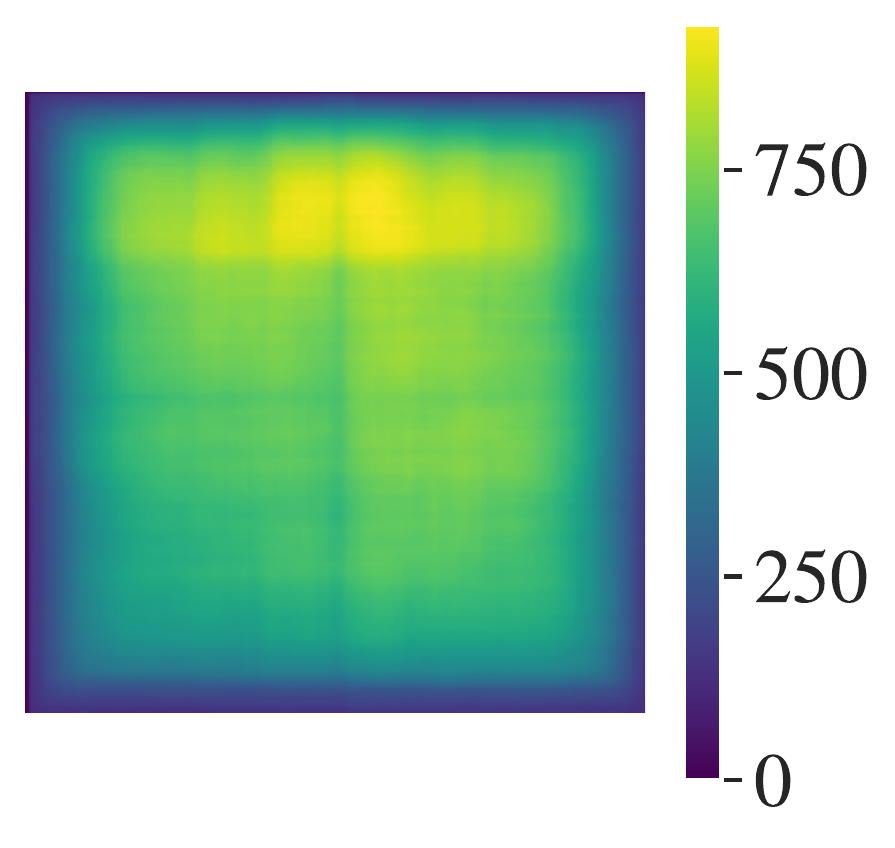}
         \caption*{Images}
     \end{subfigure}
    \begin{subfigure}[b]{0.24\textwidth}
         \centering
         \includegraphics[width=\textwidth]{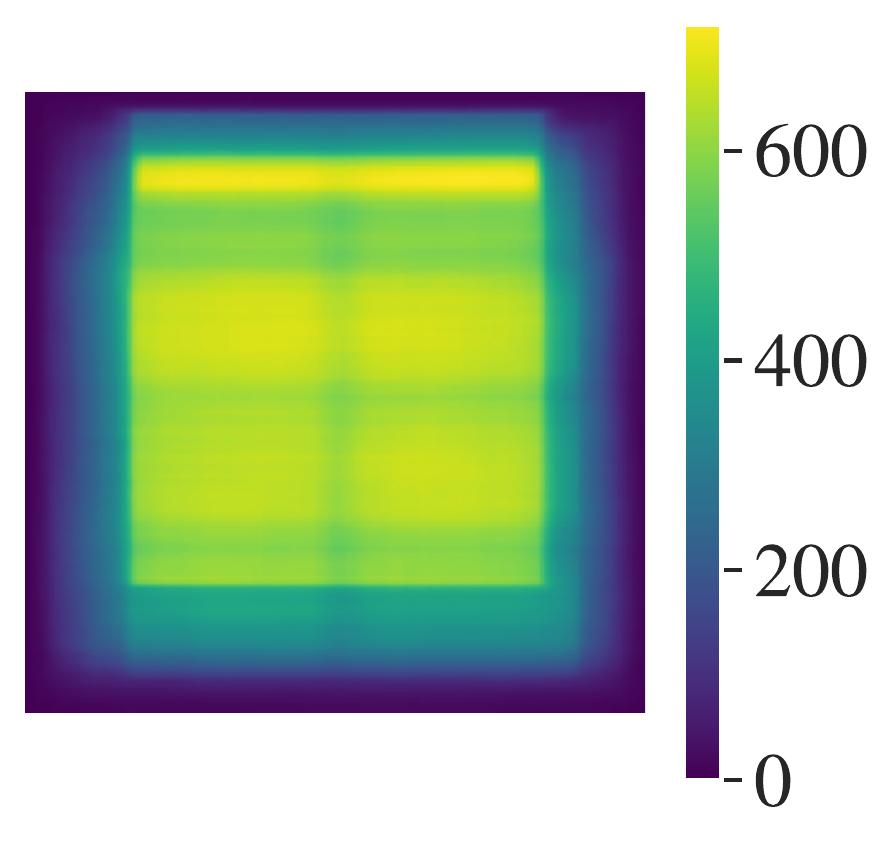}
         \caption*{Tables}
     \end{subfigure}
    \begin{subfigure}[b]{0.24\textwidth}
         \centering
         \includegraphics[width=\textwidth]{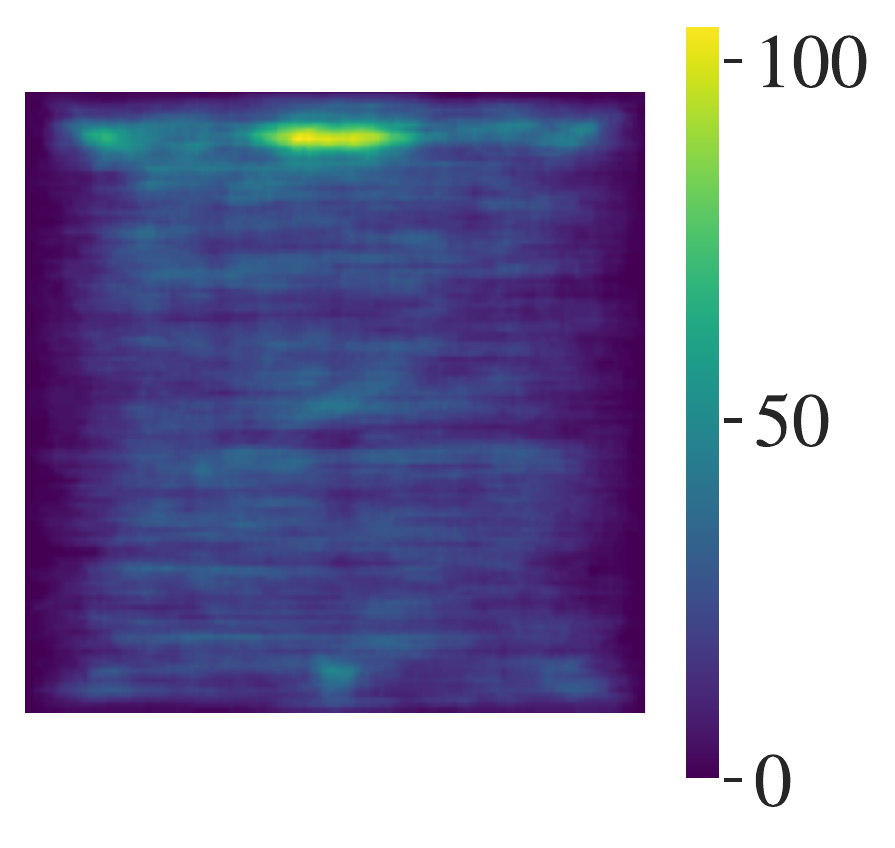}
         \caption*{Text-boxes}
     \end{subfigure}
    \begin{subfigure}[b]{0.24\textwidth}
         \centering
         \includegraphics[width=\textwidth]{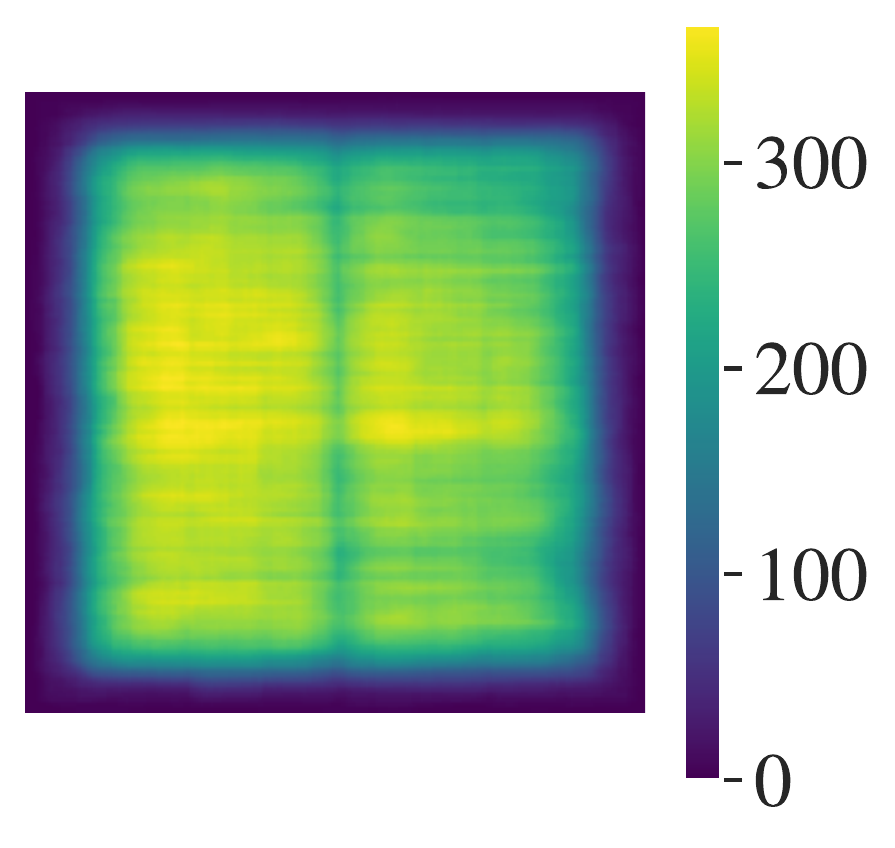}
         \caption*{Paragraphs}
     \end{subfigure}
    \caption{Un-normalized spatial distribution of annotations for different unit types. Each sample from BaDLAD is resized to a square 128x128 image, and pixel-wise density for each annotation type is presented. While for all cases there is uniformity in spatial distribution, for Text-box annotations, we see a spike in the distribution around the top, indicating high density of headers annotated as text-boxes.}
    \label{fig:heatmaps}
\end{figure}

In Fig.~\ref{fig:heatmaps}, we present the spatial distribution of different unit types for the whole dataset. The table polygons exhibit a highly concentrated localization within a distinct square shape, which is a result of the tendency for placing tables away from the borders of the document.
The text-boxes are less overlapped, as is visible in the distribution, with the exception of the header section. Conversely, paragraphs are distributed evenly throughout the body of the page and exhibit a characteristic horizontal dark bar in the center, indicative of the presence of a significant number of double-page layouts within the original dataset. 
We generate the spatial distribution by resizing each image to a 128x128 square and counting for every pixel, the number of annotations for each unit type. For the Images and Tables unit types we use all the samples from the dataset. For the text-box and paragraphs unit types, we randomly sample 50K annotations for each, to generate the figures. 


\section{Benchmark}
\label{sec:bench}
In this section, we evaluate the performance of object detection and segmentation models that are prevalent in DLA literature. We detail our methodology for generating a standard training and testing split from our dataset, report performance of benchmark models and show prediction results with qualitative analysis.

\subsection{Dataset Split}
\label{subsec:split}
The dataset was split into a train and test partition to perform our benchmarks. The split was done in a stratified method where a 60:40 train-test ratio was maintained for each domain listed in section \ref{section:distribution} except for property deed which was kept entirely in the test set. Also we ensured that the pages coming from the same book was kept in the same split to prevent data leakage. 
Previously as the authors of PubLaynet \cite{zhong2019publaynet} and HJDataset \cite{li2020docbank} claimed that segmentation masks are the quadrilateral regions for each block, Compared to the rectangular bounding boxes, they delineate the text region more accurately. The resulting train and test set had 20,365 and 13,328 samples respectively. Brief statistics of the train and test split can be found in Tables \ref{tab:domainStat} and \ref{tab:domainStatTest}.


\begin{table}[!ht]
\resizebox{\columnwidth}{!}{%
\newcolumntype{g}{>{\columncolor{Gray}}c}
\begin{tabular}{llcggggccccgggg}
    \arrayrulecolor{black}\hline
    \multirow{2}{*}{Arch.} & \multirow{2}{*}{Pretrain.} & \multirow{2}{*}{Annot.}
    & \multicolumn{4}{g}{Historical Newspapers}
    & \multicolumn{4}{c}{New Newspapers}
    & \multicolumn{4}{g}{Government Documents}\\
    & & 
    & P & Tx & I & Tb
    & P & Tx & I & Tb
    & P & Tx & I & Tb\\
    \arrayrulecolor{black}\hline
    F-RCNN    & ImgNet & BBox 
    & 57.87 & 17.49 & 59.05 & 0.0  
    & 39.08 & 12.47 & 47.60 & 2.08  
    & 43.96 & 18.68 & 22.64 & 10.70 \\
    F-RCNN    & PLNet  & BBox 
    & 64.94 & 22.10 & 67.96 & 2.38 
    & 46.74 & 16.15 & 60.68 & 14.70 
    & 46.95 & 20.03 & 28.47 & 64.35 \\
    YOLOv8    & COCO   & BBox 
    & \textbf{97.50} & \textbf{73.30} & \textbf{91.50} & \textbf{45.50} 
    & \textbf{79.70} & \textbf{45.10} & \textbf{87.50} & \textbf{64.90} 
    & \textbf{85.10} & \textbf{82.60} & \textbf{85.70} & \textbf{98.70} \\ \arrayrulecolor{gray}\hline
    F-RCNN    & ImgNet & Mask 
    & 58.30 & 18.68 & 59.59 & 0.0  
    & 40.92 & 13.46 & 47.34 & 7.29  
    & 37.72 & 18.87 & 20.48 & 7.00  \\
    M-RCNN    & ImgNet & Mask 
    & 60.33 & 18.29 & 57.30 & 0.0  
    & 41.39 & 13.15 & 45.22 & 1.91  
    & 39.06 & 18.73 & 19.43 & 3.73  \\
    M-RCNN* & PLNet  & Mask 
    & \textbf{68.63} & 22.34 & 64.08 & 3.67 
    & 48.06 & 17.12 & 55.56 & 20.21 
    & 41.57 & 21.43 & 25.88 & 49.68\\
    YOLOv8    & COCO   & Mask 
    & 64.40         & \textbf{27.10} & \textbf{77.20} & \textbf{10.80} 
    & \textbf{55.0} & \textbf{18.0} & \textbf{64.80} & \textbf{14.20} 
    & \textbf{54.90} & \textbf{22.10} & \textbf{34.80} & \textbf{45.00} \\
    \arrayrulecolor{black}\hline
    \arrayrulecolor{black}\hline
    \multirow{2}{*}{Arch.} & \multirow{2}{*}{Pretrain.}  & \multirow{2}{*}{Annot.}
    & \multicolumn{4}{g}{Magazine and Books}
    & \multicolumn{4}{c}{Liberation War Documents}
    & \multicolumn{4}{g}{Property Deeds}\\
    & &
    & P & Tx & I & Tb
    & P & Tx & I & Tb
    & P & Tx & I & Tb  \\
    \arrayrulecolor{black}\hline
    F-RCNN    & ImgNet & BBox 
    & 65.16 & 24.71 & 48.11 & 1.61  
    & 78.60 & 26.37 & 1.83 & 36.20 
    & 0.54 & 0.55 & 1.61 & 0.58\\
    F-RCNN    & PLNet  & BBox 
    & 68.91 & 26.29 & 58.36 & 15.61 
    & 79.63 & 27.64 & 1.00  & 69.60 
    & 0.34  & 0.82  & 1.24  & 0.95\\
    YOLOv8    & COCO   & BBox 
    & \textbf{93.90} & \textbf{68.20} & \textbf{79.40} & \textbf{35.00} 
    & \textbf{98.70} & \textbf{78.50} & \textbf{5.53} & \textbf{91.30} 
    & \textbf{58.00} & \textbf{51.90} & \textbf{36.40} & \textbf{57.00}\\
    \arrayrulecolor{gray}\hline
    F-RCNN    & ImgNet & Mask 
    & 61.59 & 25.52 & 46.81 & 2.86 
    & 70.40 & 26.63 & 1.34 & 40.94 
    & 0.70 & 0.58 & 1.05 & 0.86\\
    M-RCNN    & ImgNet & Mask 
    & 61.76 & 25.34 & 44.90 & 2.27 
    & 71.15 & 26.80 & 0.98  & 40.13 
    & 0.60  & 0.66  & 2.08  & 0.61\\
    M-RCNN* & PLNet  & Mask 
    & 65.77 & 27.24 & 52.03 & 11.96 
    & 72.36 & \textbf{28.87} & \textbf{2.11}  & \textbf{66.21}
    & 0.51  & 0.69  & 2.73  & 5.68\\
    YOLOv8    & COCO   & Mask 
    & \textbf{65.20} & \textbf{23.80} & \textbf{58.30} & \textbf{12.20} 
    & \textbf{72.90} & 24.50 & 1.50 & 29.70
    & \textbf{38.70} & \textbf{16.20} & \textbf{19.10} & \textbf{6.27}\\
    \arrayrulecolor{black}\hline
\end{tabular}%
}

\caption{Comparison of mAP (50-95) for different DLA architectures on BaDLAD. Models are pre-trained on ImageNet (ImgNet), PubLayNet (PLNet), and COCO datasets. We present domainwise results for each unit type, categorized as P (Paragraph), Tx (Textbox), I (Image), and Tb (Table). M-RCNN has a ResNet50 backbone whereas, M-RCNN* has a ResNet101 backbone.}

\label{tab:performance-analysis}
\end{table}

\vspace{-2em}
\subsection{Model Description and Results}

In this section we compare the performance of state-of-the-art DLA and object recognition methods on our dataset. We trained an F-RCNN and an M-RCNN model on BaDLAD utilizing the Detectron \cite{Detectron2018} implementation and a YOLOv8 model utilizing the Ultralytics \cite{jocher2020ultralytics} implementation. The R-CNN models were trained for 10,000 iterations with default hyperparameters; a learning rate of 0.001 with a decay of 0.1, a minibatch size of 48, and a warm-up iteration of 5. The YOLO models were trained for 100 epochs. For this, we used a batch size of 8, an initial learning rate of 0.01, a weight decay of 0.0005, and a warm-up iteration of 3. As a feature extractor, the RCNN models employ a ResNet-50 model, except for the M-RCNN pretrained on PubLayNet, which employs ResNet101. The performance of the benchmark models utilized in this study is presented in Table~\ref{tab:performance-analysis}. Note that while BaDLAD was labeled via polygon annotations, we also provide best fit bounding box, and segmentation masks as annotation. Therefore, models trained with an object detection target used bounding box as the ground truth, whereas, models trained with a segmentation target were evaluated on mask annotations. In accordance with the standard established by the COCO Competition~\cite{https://doi.org/10.48550/arxiv.1405.0312}, the Mean Average Precision (mAP) was computed utilizing the intersection over union (IoU) metric for bounding boxes. 
The YOLOv8 segmentation model employs a custom CNN feature extractor CSPDarknet53, in combination with a YOLO detection backbone to achieve superior accuracy in bounding box predictions across all domains and unit types. However, when it comes to mask prediction, the M-RCNN pre-trained on PubLayNet, exhibit better performance in predicting paragraphs in historical newspapers, as well as text boxes, images, and tables in Liberation war documents. YOLOv8 outperforms M-RCNN and F-RCNN in all other cases. 
YOLOv8 obtains an average mAP of $70.46\%$ and $35.69\%$ in the object recognition and segmentation settings respectively. The M-RCNN model pre-trained on PubLayNet acquired average mAP of $32.27\%$ in the segmentation setting. The models are generally more accurate in detecting paragraphs and images than text-boxes and tables. As our dataset contains a low number of table annotations, our benchmark models seem to under-perform for that unit type. However, even after being the second most frequent unit type, the accuracy of detecting text-boxes is surprisingly low. The results show that there is major scope for improvement in the DLA tasks using our dataset.

\begin{figure}[!t]
    \centering
    \begin{subfigure}[b]{0.24\textwidth}
         \centering
         \includegraphics[width=\textwidth]{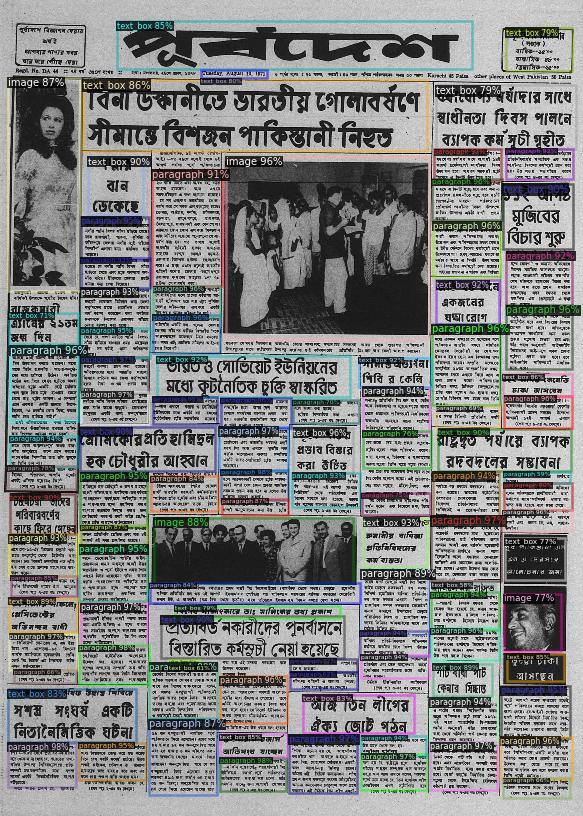}
         \caption*{}
     \end{subfigure}
    \begin{subfigure}[b]{0.24\textwidth}
         \centering
         \includegraphics[width=\textwidth]{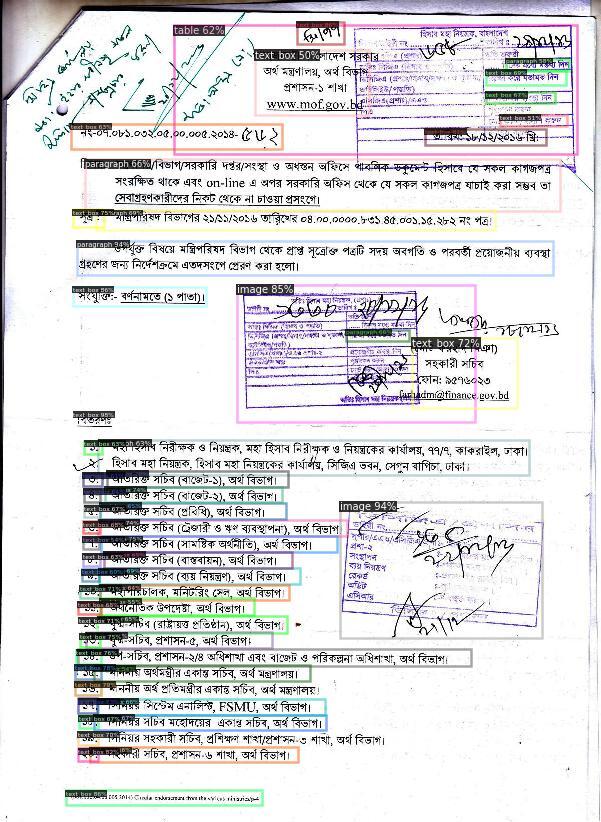}
         \caption*{}
     \end{subfigure}
    \begin{subfigure}[b]{0.24\textwidth}
         \centering
         \includegraphics[width=\textwidth]{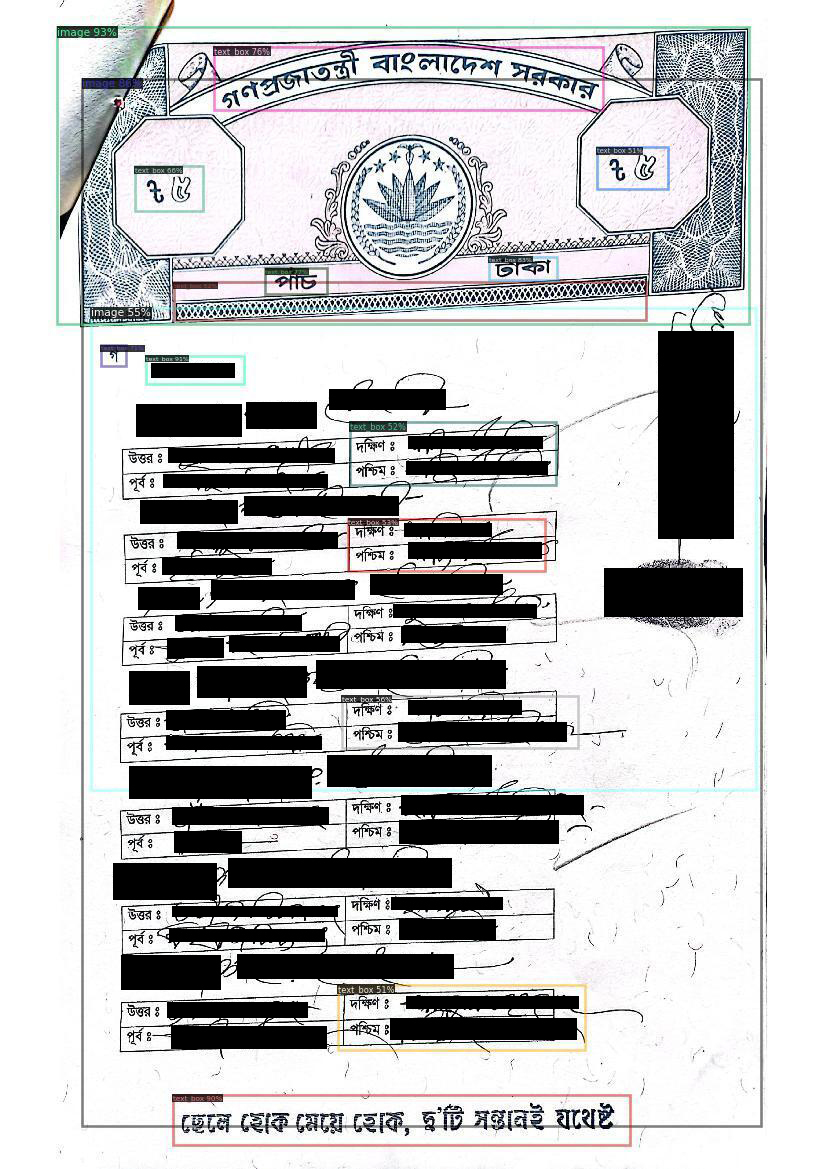}
         \caption*{}
     \end{subfigure}
    \begin{subfigure}[b]{0.24\textwidth}
         \centering
         \includegraphics[width=\textwidth]{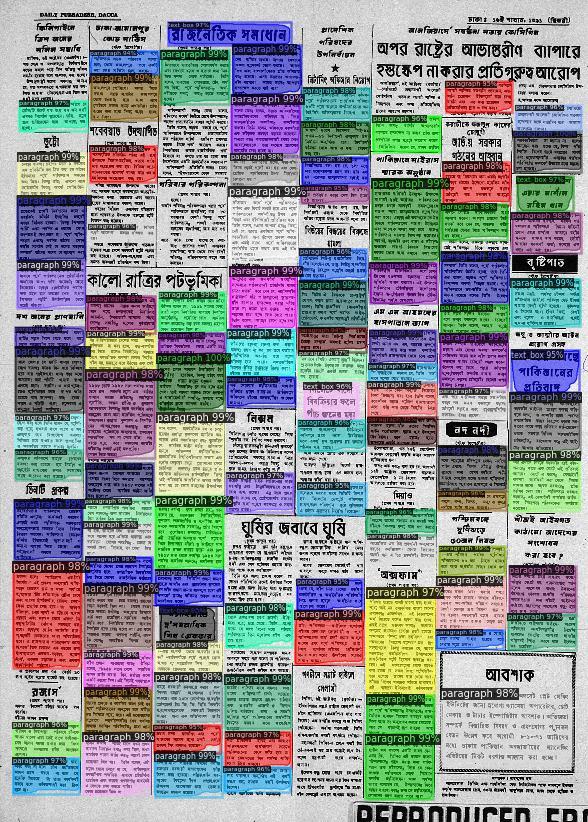}
         \caption*{}
     \end{subfigure}
    \begin{subfigure}[b]{0.20\textwidth}
         \centering
         \includegraphics[width=\textwidth]{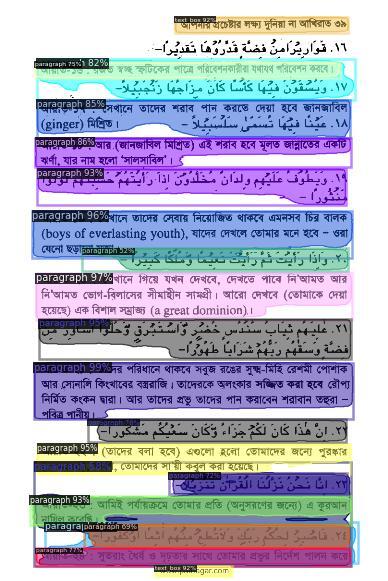}
         \caption*{}
     \end{subfigure}
    \begin{subfigure}[b]{0.35\textwidth}
         \centering
         \includegraphics[width=\textwidth]{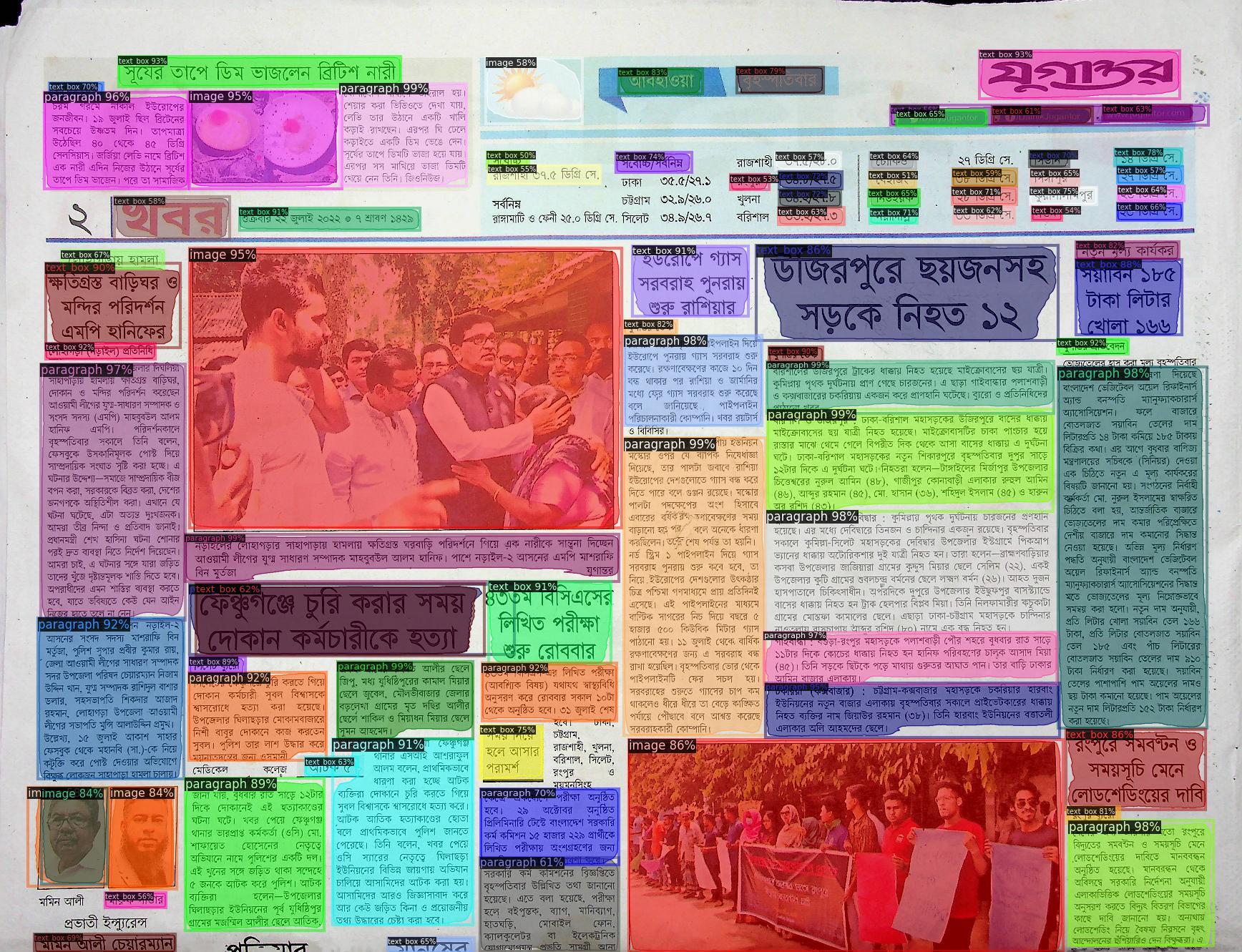}
         \caption*{}
     \end{subfigure}
    \begin{subfigure}[b]{0.40\textwidth}
         \centering
         \includegraphics[width=\textwidth]{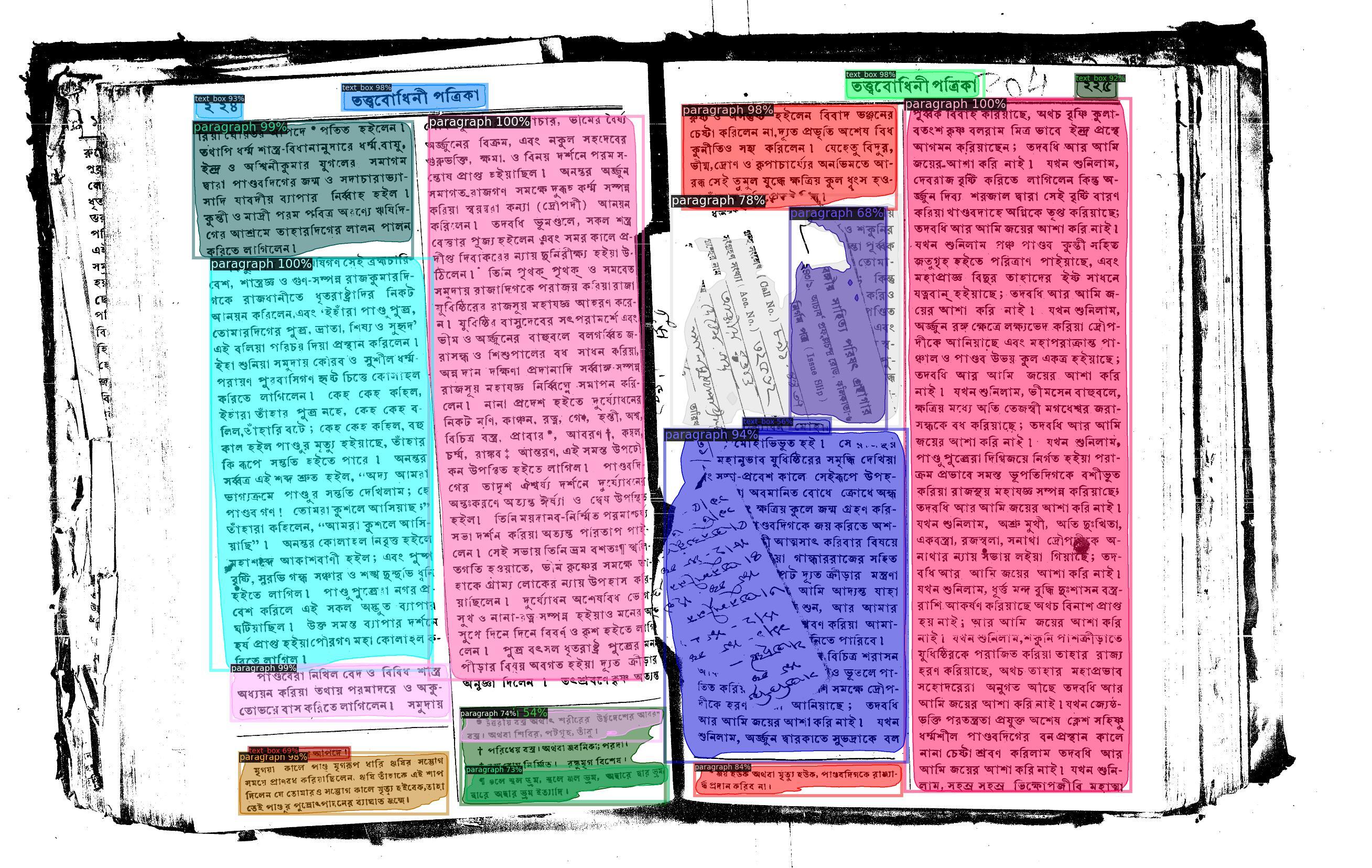}
         \caption*{}
     \end{subfigure}

    \caption{Predictions of M-RCNN-101 model on BaDLAD Test samples. The contents of the third sample (from the \textit{Property deeds} domain) has been redacted for confidentiality. The first 3 samples show only bounding box predictions and the rest show segmentation boundaries.
    }
    \label{fig:errorAnal}
\end{figure}

In Figure \ref{fig:errorAnal}, we show performance of the M-RCNN model with ResNet101 backbone, on seven test samples. For the first three samples (top-left to bottom-right) we show only the bounding box predictions while for the rest of the samples, we show both bounding box and segmentation masks predicted. 
We see that the model performs significantly bad for the third samples. This is due to the sample coming from the \textit{Property Deeds} domain which was absent in training. The fourth sample also contains a number of paragraphs which are not correctly detected, possibly due to the boldface headers. In the fifth sample, we can see the network perform well even in the presence of code-switched text. In sample six, we see the model perform very well, especially since the sample is axes aligned. For sample seven, the network exhibits robustness to noisy images and partially torn pages in the scanned document.
\section{Conclusion}
\label{sec:conclusion}

In this paper, we introduced the BaDLAD Dataset on Bengali document layout analysis and presented preliminary benchmarking results using RCNN-based and Yolo-based approaches. Unlike many prominent datasets from this domain, this is a human-annotated large dataset for layout analysis and presents an unique set of challenges. As the creation of synthetic layout analysis datasets are challenging for Bengali, this work can serve as a foundation to the field of Bengali Optical Character Recognition and also digitization of historical documents. As we are also releasing 4 million unannotated samples along with the dataset, future work can focus on utilizing unsupervised methods for training better models.

Although the dataset has diversity, it is imbalanced both in the source domains and the semantic units. This dataset will be a stepping stone in analyzing this imbalance. It can also be used as a fine-tuning dataset for a pretrained model. There are missing domains such as, shopping receipts, application form, id card etc. These domains can be added in future iterations of the dataset. We will utilize active learning methods for annotating more domain-diversified samples un-annotated portion of the dataset. One current trend in the development of DLA datasets is the use of having textual information along with layout information for Language Model based layout segmentation modeling  \cite{li2020docbank}. It is possible to get word segmentation by using a word detection algorithm \cite{ppocr} and use a word recognition model to detect the text content. We leave this as a future work, which can convert the current dataset from a segmentation one to a LM based layout analysis dataset and hence, improve the quality of segmentation performance.

\section{Acknowledgement}

We are thankful to Center for Bangladesh Genocide Research - CBGR\footnote{https://www.cbgr1971.org/} for sharing some invaluable historical documents for this dataset. We also thank the Department of Software Engineering in Shahjalal University of Science and Technology, for their support.

{
\small
\bibliographystyle{splncs04}
\bibliography{egbib}
}
\end{document}